\documentclass{article} 
\usepackage{iclr2025_conference,times}

\usepackage{amsmath,amsfonts,bm}

\def\eqref#1{equation~\ref{#1}}

\def\1{\bm{1}}

\DeclareMathAlphabet{\mathsfit}{\encodingdefault}{\sfdefault}{m}{sl}
\SetMathAlphabet{\mathsfit}{bold}{\encodingdefault}{\sfdefault}{bx}{n}

\usepackage{hyperref}
\usepackage{url}
\usepackage{bm}
\usepackage{algorithm}
\usepackage{algorithmic}
\usepackage{subfigure}
\usepackage{graphicx}
\usepackage{booktabs}

\usepackage{amsmath}
\usepackage{amssymb}
\usepackage{mathtools}
\usepackage{amsthm}

\title{Reinforced In-Context Black-Box \\Optimization}

\author{
Lei Song\thanks{Equal Contribution}$^{*1,2}$,
Chen-Xiao Gao$^{*1,2}$,
Ke Xue$^{1,2}$,
Chenyang Wu$^{1,2}$,
Dong Li$^{3}$,\\
~\textbf{Jianye Hao}$^{3,4}$\textbf{,}
\textbf{Zongzhang Zhang}$^{1,2}$\textbf{,}
\textbf{Chao Qian}\thanks{Corresponding Author}~~$^{1,2}$\\
$^1$National Key Laboratory for Novel Software Technology, Nanjing University, China\\
$^2$School of Artificial Intelligence, Nanjing University, China\\
$^3$Huawei Noah’s Ark Lab, China\\
$^4$School of Computing and Intelligence, Tianjin University, China\\
\texttt{qianc@nju.edu.cn}
}

\iclrfinalcopy 
\begin{document}

\maketitle

\begin{abstract}
Black-Box Optimization (BBO) has found successful applications in many fields of science and engineering. Recently, there has been a growing interest in meta-learning particular components of BBO algorithms to speed up optimization and get rid of tedious hand-crafted heuristics. As an extension, learning the entire algorithm from data requires the least labor from experts and can provide the most flexibility. In this paper, we propose RIBBO, a method to reinforce-learn a BBO algorithm from offline data in an end-to-end fashion. RIBBO employs expressive sequence models to learn the optimization histories produced by multiple behavior algorithms and tasks, leveraging the in-context learning ability of large models to extract task information and make decisions accordingly. Central to our method is to augment the optimization histories with \textit{regret-to-go} tokens, which are designed to represent the performance of an algorithm based on cumulative regret over the future part of the histories. The integration of regret-to-go tokens enables RIBBO to automatically generate sequences of query points that satisfy the user-desired regret, which is verified by its universally good empirical performance on diverse problems, including BBO benchmark functions, hyper-parameter optimization and robot control problems.
\end{abstract}

\section{Introduction}

Black-Box Optimization (BBO)~\cite{alarie2021two,audet2017derivative} refers to optimizing objective functions where neither analytic expressions nor derivatives of the objective are available. To solve BBO problems, we can only access the results of objective evaluation, which usually also incurs a high computational cost. Many fundamental problems in science and engineering involve optimization of expensive BBO functions, such as drug discovery~\cite{Negoescu2011TheKA,terayama2021black}, material design~\cite{frazier2015bayesian,gomez2018automatic}, robot control~\cite{calandra2016bayesian,8944013}, and optimal experimental design~\cite{greenhill2020bayesian,nguyen2023expt}, just to name a few. 

To date, a lot of BBO algorithms have been developed, among which the most prominent ones are Bayesian Optimization (BO)~\cite{bosurvey2,bosurvey1} and Evolutionary Algorithms (EA)~\cite{back:96,zhou2019evolutionary}. Despite the advancements, these algorithms typically solve BBO problems from scratch and rely on expert-derived heuristics. Consequently, they are often hindered by slow convergence rates, and unable to leverage the inherent structures within the optimization problems~\cite{10.5555/3522802.3522804,bai2023transfer}.

Recently, there has been a growing interest in meta-learning a particular component of the algorithms with previously collected data~\cite{arango2021hpob,OpenMLPython2019}. Learning the component not only alleviates the need for the laborious design process of the domain experts, but also specifies the component with domain data to facilitate subsequent optimization. For example, some components in BO are proposed to be learned from data, including the surrogate model~\cite{sam_at_icml23,NEURIPS2018_14c879f3,wang2021pre,wistuba2021fewshot}, acquisition function~\cite{hsieh2021reinforced,Volpp2020Meta-Learning}, initialization strategy~\cite{Feurer_Springenberg_Hutter_2015,poloczek2016warm}, and search space~\cite{li2022transfer,perrone2019learning}; some core evolutionary operations in EA have also been considered, e.g., learning the selection and mutation rate adaptation in genetic algorithm~\cite{lange2023discoveringga} or the update rules for evolution strategy~\cite{lange2023discoveringes}; the configuration of the algorithm can also be learned and dynamically adjusted throughout the optimization process~\cite{DBLP:conf/ecai/BiedenkappBEHL20,DBLP:journals/jair/AdriaensenBSAEL22}.

There have also been some attempts to learn an entire algorithm in an End-to-End (E2E) fashion, which requires almost no expert knowledge at all and provides the most flexibility across a broad range of BBO problems. 
However, existing practices require additional knowledge regarding the objective function during the training stage, e.g., the gradient information (often impractical for BBO)~\cite{chen2017learning,tv2019meta} or online sampling from the objective function (often very expensive)~\cite{maraval2023endtoend}.~\citet{chen2022towards} proposed the OptFormer method to imitate the behavior algorithms separately during training, presenting a challenge for the user to manually specify which algorithm to execute during testing. Thus, these methods are less ideal for practical scenarios where offline datasets are often available beforehand and a suitable algorithm for the given task has to be identified automatically without the involvement of domain experts. 



In this paper, we introduce Reinforced In-context BBO (RIBBO), which learns a reinforced BBO algorithm from offline datasets in an E2E fashion. RIBBO employs an expressive sequence model, i.e., causal transformer, to fit the optimization histories in the offline datasets generated by executing diverse behavior algorithms on multiple tasks. The sequence model is fed with previous query points and their function values, and trained to predict the distribution over the next query point. During testing, the sequence model itself serves as a BBO algorithm by generating the next query points auto-regressively. Apart from this, RIBBO augments the optimization histories with \textit{regret-to-go} (RTG) tokens, which are calculated by summing up the regrets over the future part of the histories, representing the future performance of an algorithm. A novel Hindsight Regret Relabelling (HRR) strategy is proposed to update the RTG tokens during testing. By integrating the RTG tokens into the modeling, RIBBO can automatically identify different algorithms, and generate sequences of query points that satisfy the specified regret. Such modeling enables RIBBO to circumvent the impact of inferior data and further reinforce its performance on top of the behavior algorithms.

We perform experiments on BBOB synthetic functions, hyper-parameter optimization and robot control problems by using some representatives of heuristic search, EA, and BO as behavior algorithms to generate the offline datasets. The results show that RIBBO can automatically generate sequences of query points satisfying the user-desired regret across diverse problems, and achieve good performance universally. Note that the best behavior algorithm depends on the problem at hand, and RIBBO can perform even better on some problems. Compared to the most related method OptFormer~\cite{chen2022towards}, RIBBO also has clear advantage. In addition, we perform a series of experiments to analyze the influence of important components of RIBBO.



\section{Background}

\subsection{Black-Box Optimization}\label{sec-BBO}

Let $f: \mathcal X \rightarrow \mathbb R$ be a black-box function, where $\mathcal X \subseteq \mathbb R^d$ is a $d$-dimensional search space. The goal of BBO is to find an optimal solution $\bm x^*\in \mathop{\arg\max}_{\bm x\in \mathcal X} f(\bm x)$, with the only permission of querying the objective function value. Several classes of BBO algorithms have been proposed, e.g., BO~\cite{bosurvey2,bosurvey1} and EA~\cite{back:96,zhou2019evolutionary}. The basic framework of BO contains two critical components: a surrogate model, typically formalized as Gaussian Process (GP)~\cite{gpml}, and an acquisition function~\cite{wilson2018maximizing}, which are used to model $f$ and decide the next query point, respectively. EA is a class of heuristic optimization algorithms inspired by natural evolution. It maintains a population of solutions and iterates through mutation, crossover, and selection operations to find better solutions.

To evaluate the performance of BBO algorithms, regrets are often used. The instantaneous regret $r_t=f(\bm{x}^*)-f(\bm{x}_t)$ measures the gap of function values between an optimal solution $\bm{x}^*$ and the currently selected point $\bm{x}_t$. The cumulative regret ${\rm Reg}_T=\sum_{i=1}^{T}r_i$ is the sum of instantaneous regrets in the first $T$ iterations.

\subsection{Meta-Learning in Black-Box Optimization}

Hand-crafted BBO algorithms usually require an expert to analyze the algorithms' behavior across a wide range of problems, a process that is both tedious and time-consuming. One solution is meta-learning~\cite{vilalta2002perspective,hospedales2021meta}, which aims to exploit knowledge to improve the performance of learning algorithms given data from a collection of tasks. By parameterizing a component of BBO algorithms or even an entire BBO algorithm that is traditionally manually designed, we can utilize historical data to incorporate the domain knowledge into the optimization, which may bring speedup. 

\textbf{Meta-learning particular components} has been studied with different BBO algorithms. Meta-learning in BO can be divided into four main categories according to ``what to transfer''~\cite{bai2023transfer}, including the design of the surrogate model, acquisition function, initialization strategy, and search space. For surrogate model design,~\citet{wang2021pre} and~\citet{wistuba2021fewshot} parameterized the mean or kernel function of the GP model with Multi-Layer Perceptron (MLP), while~\citet{NEURIPS2018_14c879f3} and~\citet{sam_at_icml23} substituted GP with Bayesian linear regression or neural process~\cite{garnelo2018conditional,nokey}. For acquisition function design, MetaBO~\cite{Volpp2020Meta-Learning} uses Reinforcement Learning (RL) to meta-train an acquisition function on a set of related tasks, and FSAF~\cite{hsieh2021reinforced} employs a Bayesian variant of deep Q-network as a surrogate differentiable acquisition function trained by model-agnostic meta-learning~\cite{finn2017model}. The remaining two categories focus on exploiting the previous good solutions to warm start the optimization~\cite{Feurer_Springenberg_Hutter_2015,poloczek2016warm} or shrink the search space~\cite{perrone2019learning,li2022transfer}. Meta-learning in EA usually focuses on learning specific evolutionary operations. For example, Lang et al. substituted core genetic operators, i.e., selection and mutation rate adaptation, with dot-product attention modules~\cite{lange2023discoveringga}, and meta-learned a self-attention-based architecture to discover effective and order-invariant update rules~\cite{lange2023discoveringes}. ALDes~\cite{DBLP:journals/corr/abs-2405-03419} introduces an auto-regressive learning-based approach to sequentially generate components of meta-heuristic algorithms. Beyond that, Dynamic Algorithm Configuration (DAC)~\cite{DBLP:conf/ecai/BiedenkappBEHL20,DBLP:journals/jair/AdriaensenBSAEL22} concentrates on learning the configurations of algorithms, employing RL to dynamically adjust the configurations during the optimization process.

\textbf{Meta-learning entire algorithms} has also been explored to obtain more flexible models. Early works~\cite{chen2017learning,tv2019meta} use Recurrent Neural Network (RNN) to meta-learn a BBO algorithm by optimizing the summed objective functions of some iterations. RNN uses its memory state to store information about history and outputs the next query point. This work assumes access to gradient information during the training phase, which is, however, usually impractical in BBO problems. 
OptFormer~\cite{chen2022towards} uses a text-based transformer framework to learn an algorithm, providing a universal E2E interface for BBO problems. It is trained to imitate different BBO algorithms across a broad range of problems, which, however, presents a challenge for the user to manually specify an algorithm for inference. Neural Acquisition Processes (NAP)~\cite{maraval2023endtoend} uses transformer to meta-learn the surrogate model and acquisition function of BO jointly. Due to the lack of labeled acquisition data, NAP uses an online RL algorithm with a supervised auxiliary loss for training, which requires online sampling from the expensive objective function and lacks efficiency. Black-box Optimization NETworks (BONET)~\cite{Krishnamoorthy2022GenerativePF} employ a transformer model to fit regret-augmented trajectories in an offline BBO scenario, where the training and testing data are from the same objective function, and a prefix sequence is required to warm up the optimization before testing. Compared to the above state-of-the-art E2E methods, we consider the meta-BBO setting, where the training datasets consist of diverse algorithms across different functions. Our approach offers the advantage of automatically identifying with RTG tokens and deploying the best-performing algorithm without requiring the user to pre-specify which algorithm to use or to provide a prefix sequence during the testing phase. It utilizes a supervised learning loss for training on a fixed offline dataset without the need for further interaction with the objective function. 

\subsection{Decision Transformer}
Transformer has emerged as a powerful architecture for sequence modeling tasks~\cite{khan2022transformers,wen2022transformers,wolf2020transformers}. A basic building block behind transformer is the self-attention mechanism~\cite{vaswani2017attention}, which captures the correlation between tokens of any pair of timesteps. As the scale of data and model increases, transformer has demonstrated the \textit{in-context learning} ability~\cite{NEURIPS2020_1457c0d6}, which refers to the capability of the model to infer the tasks at hand based on the input contexts. Decision Transformer (DT)~\cite{chen2021decision} abstracts RL as a sequence modeling problem, and introduces return-to-go tokens, 
representing the cumulative rewards over future interactions. Conditioning on return-to-go tokens enables DT to correlate the trajectories with their corresponding returns and generate future actions to achieve a user-specified return. Inspired by DT, we will treat BBO tasks as a sequence modeling problem naturally, use a causal transformer for modeling, and train it by conditioning on future regrets. Such design is expected to enable the learned model to distinguish algorithms with different performance and achieve good performance with a user-specified low regret.




\begin{figure*}[t!]
    \centering
    \includegraphics[width=1.0\linewidth]{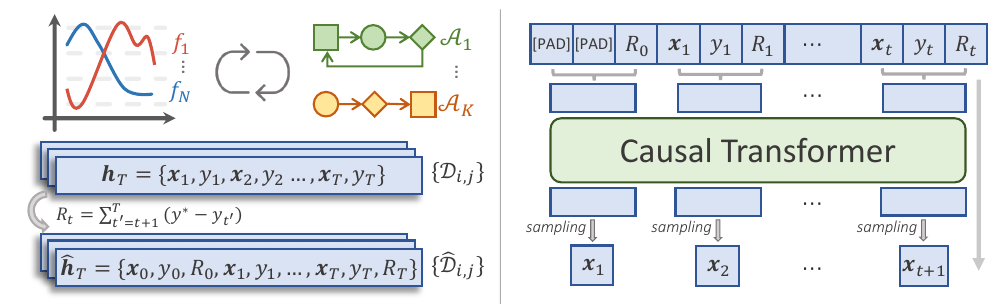}\\
    	\begin{minipage}[c]{0.45\linewidth}\centering
		\small(a) Data Generation
	\end{minipage}\ 
	\begin{minipage}[c]{0.45\linewidth}\centering
		\small(b) Training and Inference
	\end{minipage}\vspace{-0.1em}
    \caption{Illustration of RIBBO. \textit{Left: Data Generation.} $K$ existing BBO algorithms $\{\mathcal{A}_j\}_{j=1}^K$ and $N$ BBO tasks $\{f_i\}_{i=1}^N$ are used to serve as the behavior algorithms and the training tasks, respectively. The offline datasets $\{\mathcal{D}_{i,j}\}$ consist of the optimization histories $\bm h_T=\{(\bm x_t, y_t)\}_{t=1}^T$ collected by executing each behavior algorithm $\mathcal{A}_j$ on each task $f_i$ for $T$ evaluation steps, which are then augmented with the regret-to-go tokens $R_t$ (calculated as the cumulative regret $\sum^T_{t'=t+1} (y^*-y_{t'})$ over the future optimization history) to generate the final dataset $\{\widehat{\mathcal{D}}_{i,j}\}$ for training. \textit{Right: Training and Inference.} Our model takes in triplets of $(\bm{x}_t, y_t, R_t)$, embeds them into one token, and outputs the distribution over the next query point $\bm x_{t+1}$. During training, the ground-truth next query point is used to minimize the loss in Eq.~(\ref{eq:ribbo_objective}). During inference, the next query point $\bm x_{t+1}$ is generated auto-regressively based on the current history $\hat{\bm h}_t$.}
    \label{fig:banner}\vspace{-0.9em}
\end{figure*}

\section{Method}
This section presents Reinforced In-context Black-Box Optimization~(RIBBO), which learns an enhanced BBO algorithm in an E2E fashion, as illustrated in Figure~\ref{fig:banner}. We follow the task-distribution assumption, which is commonly adopted in meta-learning settings~\cite{finn2017model,hospedales2021meta,omni-vrp}. Our goal is to learn a generalizable model $\mathcal M$ capable of solving a wide range of BBO tasks, each associated with a BBO objective function $f$ sampled from the task distribution $P(\mathcal F)$, where $\mathcal F$ denotes the function space. 

Let $[N]$ denote the integer set $\{1,2,\ldots,N\}$. During training, we usually access $N$ source tasks and each task corresponds to an objective function $f_i\sim P(\mathcal F)$, where $i\in [N]$. Hereafter, we use $f_i$ to denote the task $i$ if the context is clear. We assume that the information is available via offline datasets $\mathcal D_{i, j}$, which are produced by executing a behavior algorithm $\mathcal A_j$ on task $f_i$, where $j\in [K]$ and $i\in [N]$. Each dataset $\mathcal D_{i, j} = \{\bm h^{i, j, m}_T\}_{m=1}^M$ consists of $M$ optimization histories $\bm h^{i, j, m}_T = \{(\bm x_t, y_t)\}_{t=1}^T$, where $\bm x_t$ is the query point selected by $\mathcal A_j$ at iteration $t$, and $y_t = f_i(\bm x_t)$ is its objective value. If the context is clear, we will omit $i, j, m$ and simply use $\bm h_T$ to denote a history with length $T$. The initial history $\bm h_{0}$ is defined as $\emptyset$. We impose no additional assumptions about the behavior algorithms, allowing for a range of BBO algorithms, even random search. 


With the datasets, we seek to learn a model $\mathcal{M}_{\bm \theta}(\bm x_t| \bm h_{t-1})$, which is parameterized by $\bm \theta$ and generates the next query point $\bm x_t$ by conditioning on the previous history $\bm h_{t-1}$. As introduced in Section~\ref{sec-BBO}, with a given budget $T$ and the history $\bm h_T$ produced by an algorithm $\mathcal{A}$, we use the cumulative regret
\begin{equation}
\label{eq:reg}{\rm Reg}_T =\sum_{t=1}\nolimits^T (y^*-y_t) 
\end{equation}
as the evaluation metric, where $y^*$ is the optimum value and $\{y_t\}^T_{t=1}$ are the function values in $\bm h_T$.




\subsection{Method Outline}


Given the current history $\bm h_{t-1}$ at iteration $t$, a BBO algorithm selects the next query point $\bm x_t$, observes the function value $y_t = f_i(\bm x_t)$, and updates the history $\bm h_t=\bm h_{t-1}\cup \{(\bm x_t, y_t)\}$. Similar to the previous work~\cite{chen2017learning}, we take this framework as a starting point and treat the learning of a universal BBO algorithm as learning a model $\mathcal{M}_{\bm \theta}$, which takes the preceding history $\bm h_{t-1}$ as input and outputs a distribution of the next query point $\bm x_t$. The optimization histories in offline datasets provide a natural supervision for the learning process.
 
Suppose we have a set of histories $\{\bm h_T\}$, generated by a single behavior algorithm $\mathcal A$ on a single task $f$. By employing a causal transformer model $\mathcal{M}_{\bm \theta}$, we expect $\mathcal{M}_{\bm \theta}$ to imitate $\mathcal{A}$ and produce similar optimization history on $f$. In practice, we usually have datasets containing histories from multiple behavior algorithms $\{\mathcal{A}_j\}_{j=1}^K$ on multiple tasks $\{f_i\}_{i=1}^N$. To fit $\mathcal{M}_{\bm \theta}$, we use the negative log-likelihood loss 
\begin{equation}\label{eq:bc}
    \mathcal{L}_{\rm BC}(\bm \theta)=-\mathbb{E}_{\bm h_T\sim \mathcal{D}_{i, j}}\left[\sum\nolimits_{t=1}^T \log \mathcal{M}_{\bm \theta}(\bm x_{t}|\bm h_{t-1})\right]. 
\end{equation}
To effectively minimize the loss, $\mathcal{M}_{\bm \theta}$ needs to recognize both the task and the behavior algorithm in-context, and then imitate the optimization behavior of the corresponding behavior algorithm. 

Nevertheless, naively imitating the offline datasets hinders the model since some inferior behavior algorithms may severely degenerate the model's performance. Inspired by DT~\cite{chen2021decision}, we propose to augment the optimization history with Regret-To-Go~(RTG) tokens $R_t$, defined as the sum of instantaneous regrets over the future history:
\begin{equation}\label{eq:RTG}
    \hat{\bm h}_T = \{(\bm x_t, y_t, R_t)\}_{t=0}^T, \;R_t=\sum\nolimits_{t'=t+1}^T (y^*-y_{t'}), 
\end{equation}
where $\bm x_0$ and $y_0$ are placeholders for padding, denoted as \text{[PAD]} in Figure~\ref{fig:banner}(b), and $R_T=0$. The augmented histories compose the augmented dataset $\widehat{\mathcal{D}}_{i, j}$, and the training objective of $\mathcal{M}_{\bm \theta}$ becomes
\begin{equation}\label{eq:ribbo_objective}
    \mathcal{L}_{\rm{RIBBO}}(\bm \theta)=-\mathbb{E}_{\hat{\bm h}_T\sim \widehat{\mathcal{D}}_{i, j}}\left[\sum\nolimits_{t=1}^T \log \mathcal{M}_{\bm \theta}(\bm x_{t}|\hat{\bm h}_{t-1})\right]. 
\end{equation}
The integration of RTG tokens in the context brings identifiability of behavior algorithms, and the model $\mathcal{M}_{\bm \theta}$ can effectively utilize them to make appropriate decisions. Furthermore, RTG tokens have a direct correlation with the metric of interest, i.e., cumulative regret $\mathrm{Reg}_T$ in Eq.~(\ref{eq:reg}). Conditioning on a lower RTG token provides a guidance to our model and reinforces $\mathcal{M}_{\bm \theta}$ to exhibit superior performance. These advantages will be clearly shown by experiments in Section~\ref{sec:rtg-analysis}.

The resulting method RIBBO has implicitly utilized the in-context learning capacity of transformer to guide the optimization with previous histories and the desired future regret as context. The in-context learning capacity of inferring the tasks at hand based on the input contexts has been observed as the scale of data and model increases~\cite{Kaplan2020ScalingLF}. It has been explored to infer general functional relationships as supervised learning or RL algorithms. For example, the model is expected to predict accurately on the query input $\bm x_t$ by feeding the training dataset $\{(\bm x_i, y_i)\}_{i=1}^{t-1}$ as the context~\cite{guo2023transformers,hollmann2023tabpfn,li2023transformers};~\citet{laskin2023incontext} learned RL algorithms using causal transformers. Here, we use it for BBO.


\subsection{Practical Implementation}
\label{sec-practical-imple}
Next, we detail the model architecture, training, and inference of RIBBO. 

\textbf{Model Architecture. }
For the formalization of the model $\mathcal{M}_{\bm \theta}$, we adopt the commonly used GPT architecture~\cite{radford2018improving}, which comprises a stack of causal attention blocks. Each block is composed of an attention mechanism and a feed-forward network. We aggregate each triplet $( \bm x_i, y_i, R_i )$ using a two-layer MLP network. The output of $\mathcal{M}_{\bm \theta}$ is a diagonal Gaussian distribution of the next query point. Note that previous works that adopt the sequence model as surrogate models~\cite{nokey,sam_at_icml23,Nguyen2022TransformerNP} typically remove the positional encoding because the surrogate model should be invariant to the history order. On the contrary, our implementation preserves the positional encoding, naturally following the behavior of certain algorithms (e.g., BO or EA) and making it easier to learn from algorithms. Additionally, the positional encoding can help maintain the monotonically decreasing order of RTG tokens. More details about the architecture can be found in Appendix~\ref{appendix:model-details}. 

\textbf{Model Training. }
RTG tokens are calculated as outlined in Eq.~(\refeq{eq:RTG}) for the offline datasets before training. Since the calculation of regret requires the optimum value of task $i$, we use the best-observed value $y_{\rm max}^i$ as a proxy for the optimum. Let $\{\widehat{\mathcal{D}}_{i, j}\}_{i\in [N],j\in[K]}$ denote the RTG augmented datasets with $N$ tasks and $K$ algorithms. During training, we sample a minibatch of consecutive subsequences of length $\tau<T$ uniformly from the augmented datasets. The training objective is to minimize the RIBBO loss in Eq.~(\refeq{eq:ribbo_objective}).

\begin{algorithm}[t!]
\caption{Model Inference with HRR}
\label{algo:hrr}
{\textbf{Input}:} trained model $\mathcal M_{\bm \theta}$, budget $T$, optimum value $y^*$\\
{\textbf{Process}:}
\begin{algorithmic}[1]
\STATE Initialize $\hat{\bm h}_0 = \{(\bm x_0, y_0, R_0)\}$, where $\bm x_0$ and $y_0$ are placeholders for padding and $R_0 = 0$;
\FOR {$t=1, 2, \ldots, T$}
\STATE Generate the next query point $\bm x_t \sim \mathcal M_{\bm \theta}(\bm \cdot| \hat{\bm h}_{t-1})$;
\STATE Evaluate $\bm x_t$ to obtain $y_t=f(\bm{x}_t)$;
\STATE Calculate the instantaneous regret $r = y^* - y_t$;
\STATE Relabel $R_i \gets R_i + r$, for each $(\bm{x}_i, y_i, R_i)$ in $\hat{\bm h}_{t-1}$;
\STATE $\hat{\bm h}_t = \hat{\bm h}_{t-1} \cup \{(\bm{x}_t, y_t, 0)\}$
\ENDFOR
\end{algorithmic}
\end{algorithm}

\textbf{Model Inference.} The model $\mathcal{M}_{\bm \theta}$ generates the query points $\bm x_t$ auto-regressively during inference, which involves iteratively selecting a new query point $\bm x_t$ based on the current augmented history $\hat{\bm h}_{t-1}$, evaluating the query as $y_t=f(\bm x_t)$, and updating the history by $\hat{\bm h}_t=\hat{\bm h}_{t-1}\cup \{(\bm x_t, y_t, R_t)\}$. A critical aspect of this process is how to specify the value of RTG (i.e., $R_t$) at every iteration $t$. Inspired by DT, a naive approach is to specify a desired performance as the initial RTG $R_0$, and decrease it as $R_t=R_{t-1}-(y^* - y_t)$. However, this strategy has the risk of producing out-of-distribution RTGs, since the values can fall below $0$ due to an improperly selected $R_0$. 

Given the fact that RTGs are lower bounded by $0$ and a value of $0$ implies a good BBO algorithm with low regret, we propose to set the immediate RTG as $0$. Furthermore, we introduce a strategy called \textbf{Hindsight Regret Relabelling (HRR)} to update previous RTGs based on the current sample evaluations. The inference procedure with HRR is detailed in Algorithm~\ref{algo:hrr}. In line~1, the history $\hat{\bm h}_0$ is initialized with padding placeholders $\bm x_0, y_0$ and RTG $R_0=0$. At iteration $t$ (i.e., lines~3--7), the model $\mathcal M_{\bm \theta}$ is fed with the augmented history $\hat{\bm h}_{t-1}$ to generate the next query point $\bm x_t$ in line~3, followed by the evaluation procedure to obtain $y_t$ in line~4. Then, the immediate RTG $R_t$ is set to $0$, and we employ HRR to update previous RTG tokens in $\hat{\bm h}_{t-1}$, i.e., calculate the instantaneous regret $r = y^* - y_t$ (line~5) and add $r$ to every RTG token within $\hat{\bm h}_{t-1}$ (line~6):
\begin{align}\label{eq-hrr}
\forall 0\leq i <t, R_i\gets R_i+ (y^* -y_t).
\end{align}
Note that this relabelling process guarantees that $\forall 0\leq i<t$, the RTG token $R_{i}=\sum_{t'=i+1}^t (y^* - y_{t'})$, which can also be written as $\sum_{t'=i+1}^T (y^* - y_{t'})$, consistent with the definition in Eq.~(\ref{eq:RTG}), because the immediate RTG $R_t=\sum_{t'=t+1}^T (y^* - y_{t'})$ is set to 0. In line~7, the history $\hat{\bm h}_t$ is updated by expanding $\hat{\bm h}_{t-1}$ with $\{(\bm{x}_t, y_t, 0)\}$, i.e., the current sampling and its immediate RTG $R_t=0$. The above process is repeated until reaching the budget $T$. Thus, we can find that HRR not only exploits the full potential of $\mathcal{M}_{\bm \theta}$ through using $0$ as the immediate RTG and thereby demands the model to generate the most advantageous decisions, but also preserves the calculation of RTG tokens following the same way as the training data, i.e., representing the cumulative regret over future optimization history.

\subsection{Data Generation}

Finally, we give some guidelines about data generation for using the proposed RIBBO method.

\textbf{Data Collection.} Given a set of tasks $\{f_i\}_{i=1}^N$ sampled from the task distribution $P(\mathcal F)$, we can employ a diverse set of behavior algorithms for data collection. For example, we can select some representatives from different types of BBO algorithms, e.g., BO and EA. Datasets $\mathcal D_{i, j}$ are obtained by using each behavior algorithm to optimize each task with different random seeds. Each optimization history $\bm{h}_T=\{(\bm x_t, y_t)\}_{t=1}^T$ in $\mathcal D_{i, j}$ is then augmented with RTG tokens $R_t$, which is computed as in Eq.~(\ref{eq:RTG}). The resulting histories $\hat{\bm{h}}_T=\{(\bm x_t, y_t,R_t)\}_{t=0}^T$ compose the final datasets $\widehat{\mathcal D}_{i, j}$ for model training.

\textbf{Data Normalization.}
To provide a unified interface and balance the statistic scales across tasks, it is important to apply normalization to the inputs to our model. We normalize the point $\bm x$ by $(\bm x-\bm x_{\rm min})/(\bm x_{\rm max}-\bm x_{\rm min})$, with $\bm x_{\rm max}$ and $\bm x_{\rm min}$ being the upper and lower bounds of the search space, respectively. For the function value $y$, we apply random scaling akin to previous works~\cite{wistuba2021fewshot,chen2022towards}. That is, when sampling a history $\bm h_{\tau}$ from the datasets $\mathcal D_{i, j}$, we randomly sample the lower bound $l\sim \mathcal{U}(y^i_{\rm min}-\frac s2, y^i_{\rm min}+\frac s2)$ and the upper bound $u\sim \mathcal{U}(y^i_{\rm max}-\frac s2, y^i_{\rm max}+\frac s2)$, where $\mathcal U$ stands for uniform distribution, $y^i_{\rm min}, y^i_{\rm max}$ denote the observed minimum and maximum values for $f_i$, and $s=y^i_{\rm max}-y^i_{\rm min}$; the values $y_t$ in $\bm h_{\tau}$ are then normalized by $(y_t-l)/(u-l)$ for training. The RTG tokens are calculated accordingly with the normalized values. The random normalization can make a model exhibit invariance across various scales of $y$. For inference, the average values of the best-observed and worst-observed values across the training tasks are used to normalize $y$.

\section{Experiments}
\label{sec:exp}

In this section, we examine the performance of RIBBO on a wide range of tasks, including synthetic functions, Hyper-Parameter Optimization (HPO) and robot control problems. The model architecture and hyper-parameters are maintained consistently across these problems. We train our model using five distinct random seeds, ranging from $0$ to $4$, and each trained model is run five times independently during the execution phase. We will report the average performance and standard deviation. Details of the model hyper-parameters are given in Appendix~\ref{appendix:model-details}. Our code is available at \url{https://github.com/songlei00/RIBBO}.

\subsection{Experimental Setup}

\textbf{Benchmarks.} We use BBO Benchmarks BBOB~\cite{ElHara2019COCOTL}, HPO-B~\cite{arango2021hpob}, and rover trajectory planning task~\cite{wang2018batched}. The BBOB suite, a comprehensive and widely used benchmark in the continuous domain, consists of $24$ synthetic functions. For each function, a series of linear and non-linear transformations are implemented on the search space to obtain a distribution of functions with similar properties. According to the properties of these functions, they can be divided into $5$ categories, and we select one from each category due to resource constraints, including Greiwank Rosenbrock, Lunacek, Rastrigin, Rosenbrock, and Sharp Ridge. HPO-B is a commonly used HPO benchmark and consists of a series of HPO problems. Each problem is to optimize a machine learning model across various datasets, and an XGBoost model is provided as the objective function for evaluation in a continuous space. We conduct experiments on two widely used models, SVM and XGBoost, in the continue domain. For robot control optimization, we perform experiments on rover trajectory planning task, which is a trajectory optimization problem to emulate rover navigation. Similar to~\cite{ElHara2019COCOTL,Volpp2020Meta-Learning}, we implement random translations and scalings to the search space to construct a distribution of functions. For BBOB and rover problems, we sample a set of functions from the task distribution as training and test tasks, while for HPO-B, we use the meta-training/test task splits provided by the authors. Detailed explanations of the benchmarks can be found in Appendix~\ref{appendix:benchmarks}.


\textbf{Data.} Similar to OptFormer~\cite{chen2022towards}, we employ $7$ behavior algorithms, i.e., Random Search, Shuffled Grid Search, Hill Climbing, Regularized Evolution~\cite{real2019regularized}, Eagle Strategy~\cite{yang2010eagle}, CMA-ES~\cite{cmaes}, and GP-EI~\cite{balandat2020botorch}, which are representatives of heuristic search, EA, and BO, respectively. Datasets are generated by employing each behavior algorithm to optimize various training functions sampled from the task distribution, using different random seeds. For inference, new test functions are sampled from the same task distribution to serve as the test set. Specifically, for the HPO-B problem, the meta-training/test splits have been predefined by the authors and we adhere to this standard setup. Additional information about the behavior algorithms and datasets can be found in Appendix~\ref{appendix:behavior-algo} and~\ref{appendix:data}. 

\subsection{Baselines}

As RIBBO is an in-context E2E model, the most related baselines are those also training an E2E model with offline datasets, including Behavior Cloning (BC)~\cite{bain1995framework} and OptFormer~\cite{chen2022towards}. Their hyper-parameters are set as same as that of our model for fairness. Note that the seven behavior algorithms used to generate datasets are also important baselines, and included for comparison as well. 

\textbf{BC} uses the same transformer architecture as RIBBO. The only difference is that we do not feed RTG tokens into the model of BC and train to minimize the BC loss in Eq.~(\refeq{eq:bc}). When the solutions are generated auto-regressively, BC tends to imitate the average behavior of various behavior algorithms. Consequently, the inclusion of underperforming algorithms, e.g., Random Search and Shuffled Grid Search, may significantly degrade the performance. To mitigate this issue, we have also trained the model by excluding these underperforming algorithms, denoted as \textbf{BC Filter}.

\textbf{OptFormer} employs a transformer to imitate the behaviors of a set of algorithms and an algorithm identifier usually needs to be specified manually during inference for superior performance. Its original implementation is built upon a text-based transformer with a large training scale. In this paper, we re-implement a simplified version of OptFormer where we only retain the algorithm identifier within the metadata. The initial states, denoted as $\bm x_0$ and $y_0$, are used to distinguish between algorithms. They are obtained by indexing the algorithm type through an embedding layer, thereby aligning the initial states with the specific imitated algorithm. This enables the identification of distinct behavior algorithms within the simplified OptFormer. Further details about the re-implementation can be found in Appendix~\ref{appendix:optformer}.

\subsection{Main Results}

\begin{figure*}[!t]
    \centering
    \subfigure{\includegraphics[width=0.45\textwidth]{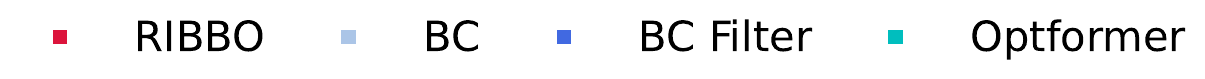}}\\\vspace{-1.2em}
    \subfigure{\includegraphics[width=0.8\textwidth]{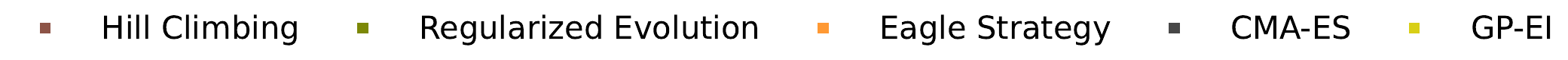}}\\\vspace{-0.9em}
    \centering
    \subfigure{\includegraphics[width=0.24\textwidth]{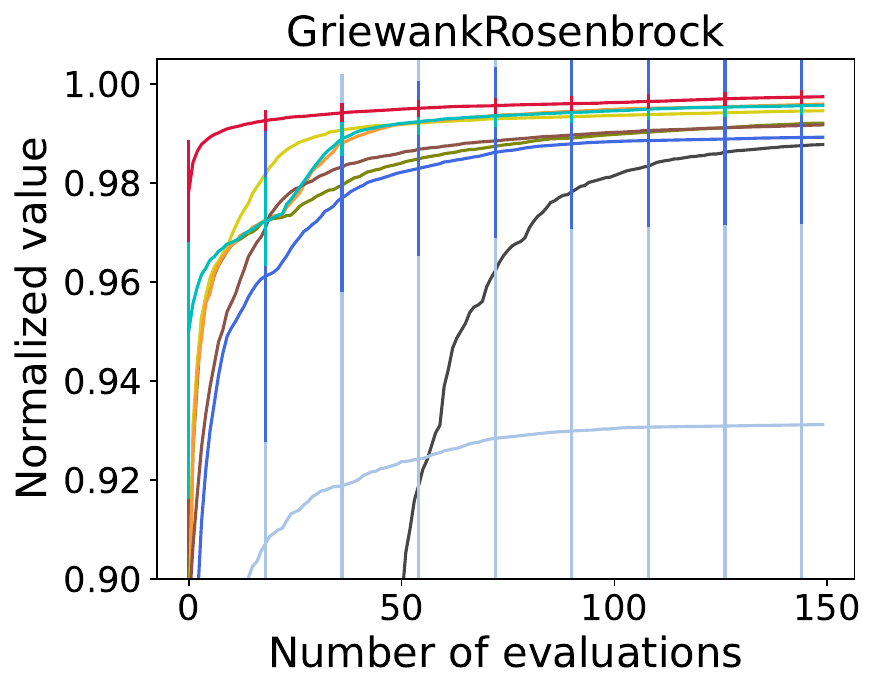}}
    \subfigure{\includegraphics[width=0.24\textwidth]{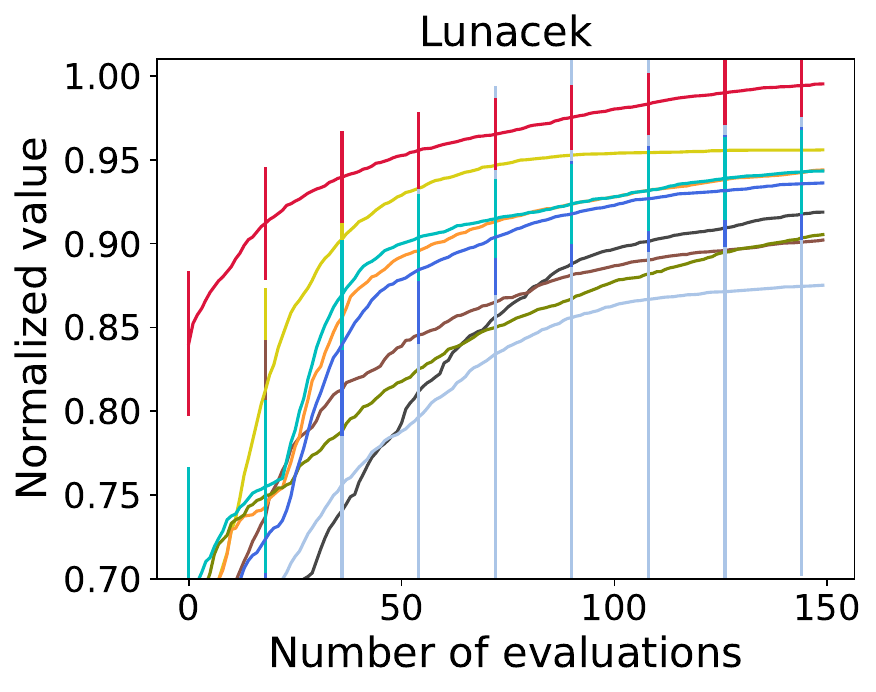}}
    \subfigure{\includegraphics[width=0.24\textwidth]{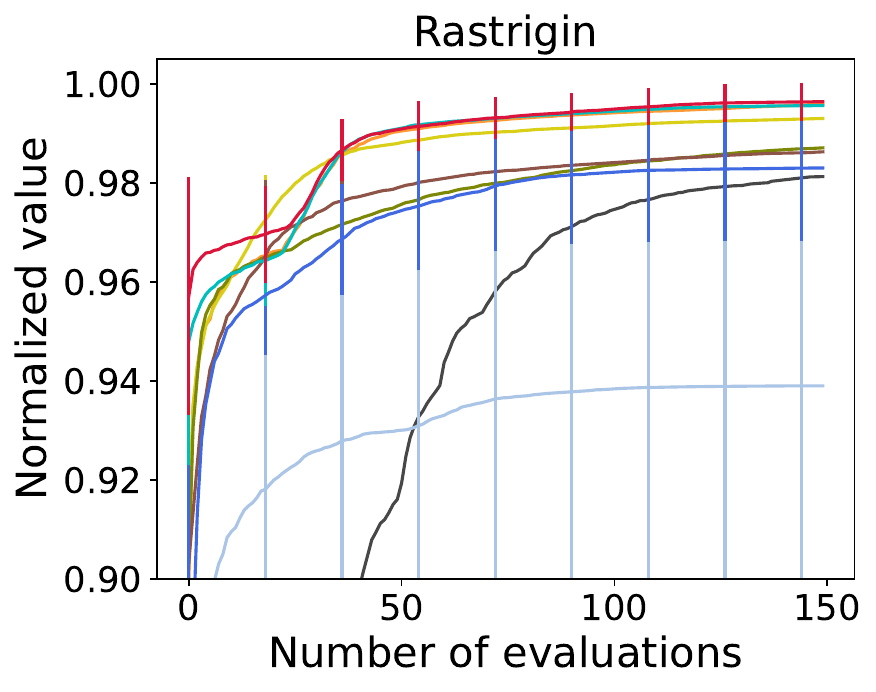}}
    \subfigure{\includegraphics[width=0.24\textwidth]{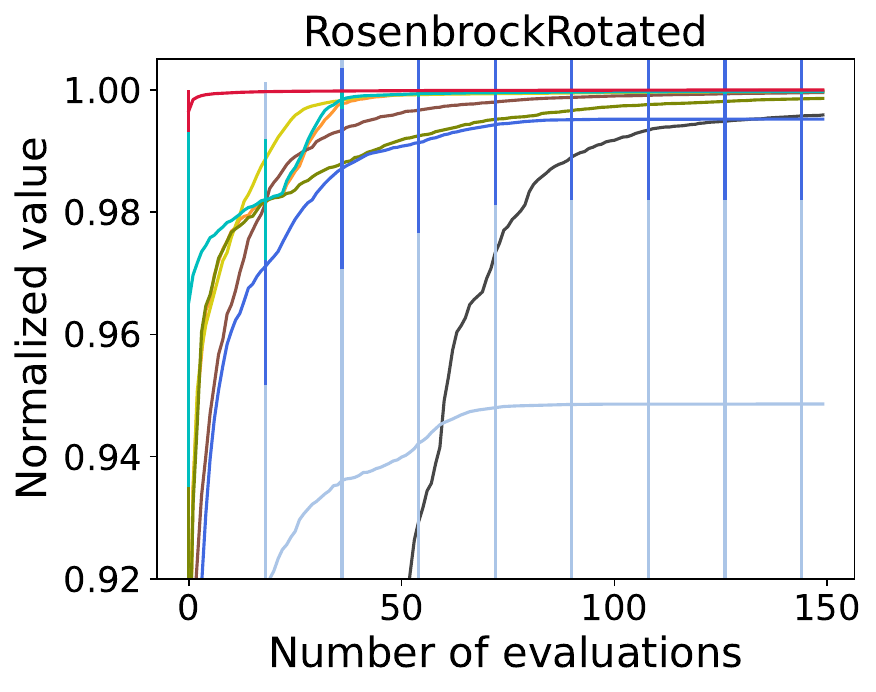}}\\\vspace{-0.8em}
    \subfigure{\includegraphics[width=0.24\textwidth]{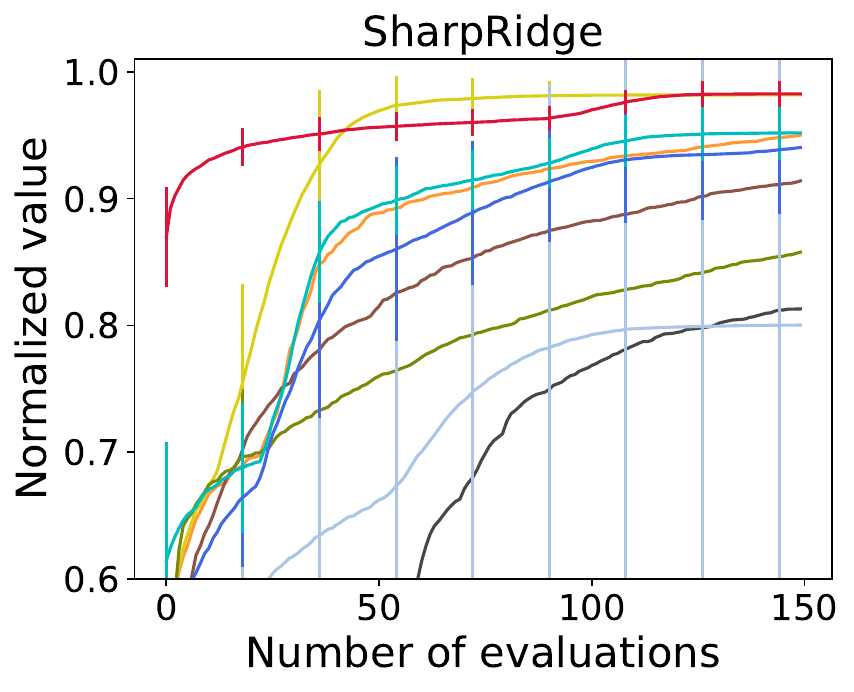}}
    \subfigure{\includegraphics[width=0.24\textwidth]{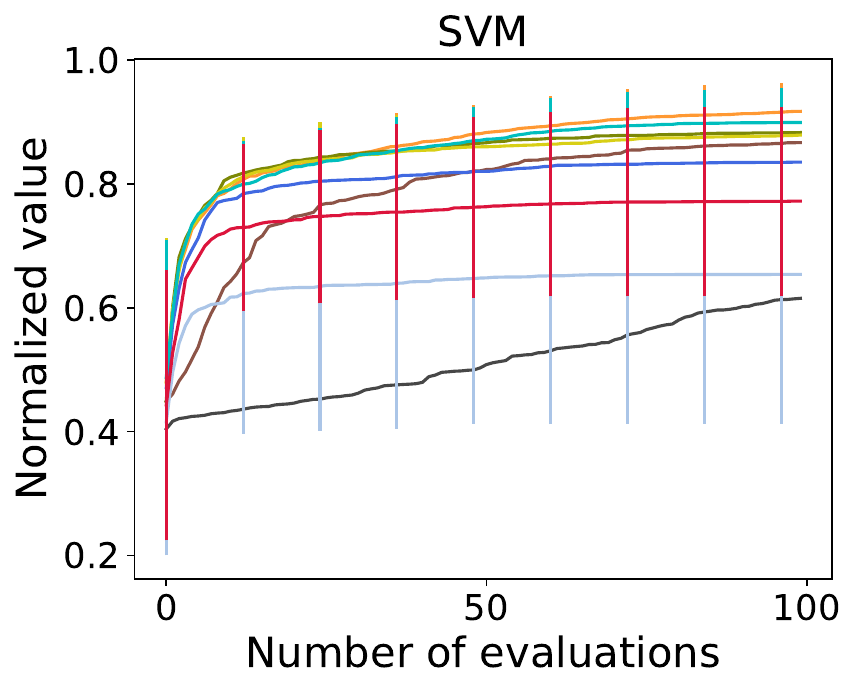}}
    \subfigure{\includegraphics[width=0.24\textwidth]{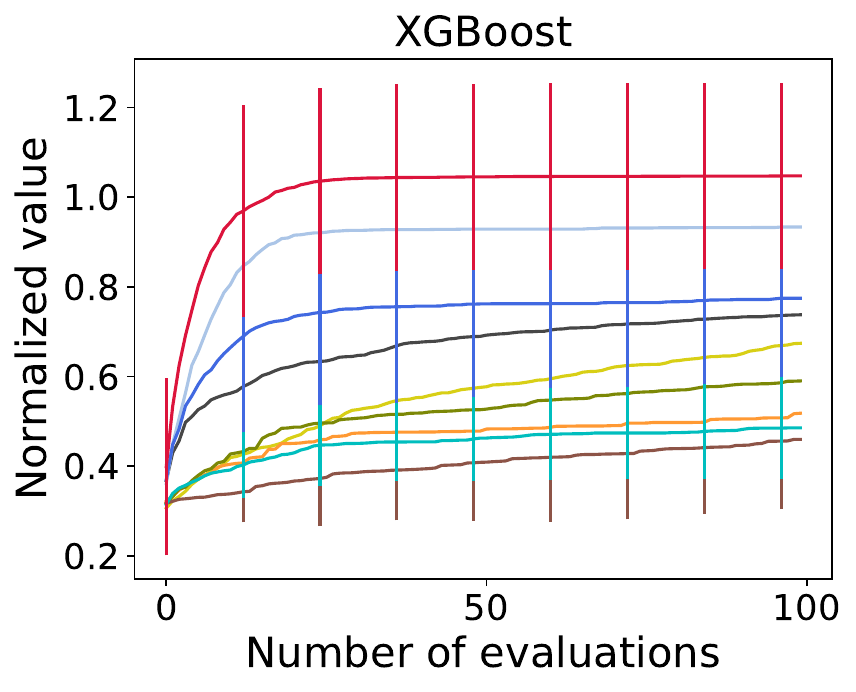}}
    \subfigure{\includegraphics[width=0.24\textwidth]{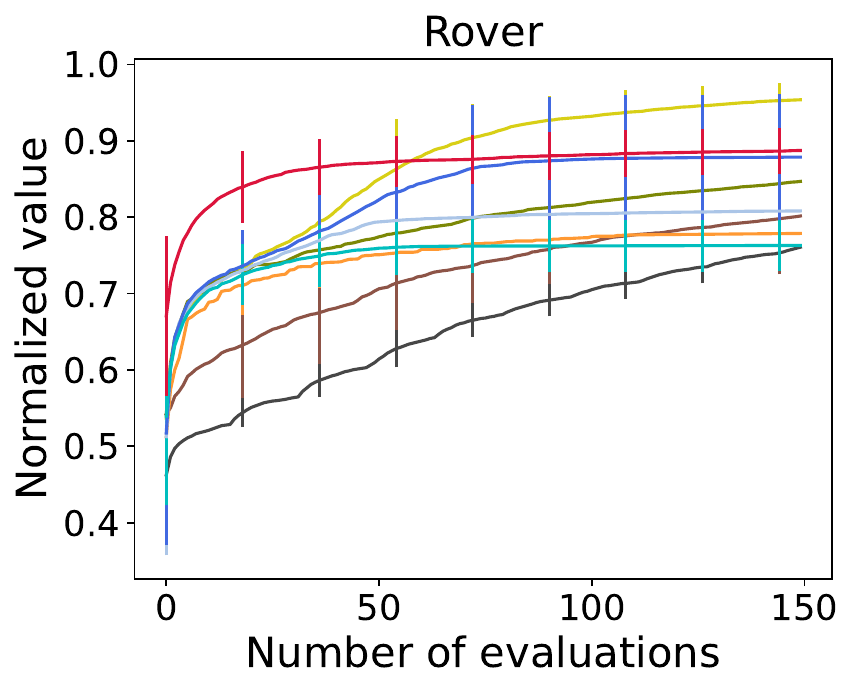}}\vspace{-0.5em}
    \caption{Performance comparison among RIBBO, BC, BC Filter, OptFormer, and behavior algorithms on synthetic functions, HPO, and robot control problems. The $y$-axis is the normalized average objective value, and the length of vertical bars represents the standard deviation.}
    \label{fig:main_exp}\vspace{-0.9em}
\end{figure*}

The results are shown in Figure~\ref{fig:main_exp}. For the sake of clarity in visualization, we have omitted the inclusion of Random Search and Shuffled Grid Search due to their poor performance from start to finish. We can observe that RIBBO achieves superior or at least equivalent efficacy in comparison to the best behavior algorithm on each problem except SVM and rover. This demonstrates the versatility of RIBBO, while the most effective behavior algorithm depends upon the specific problem at hand, e.g., the best behavior algorithms on Lunacek, Rastrigin and XGBoost are GP-EI, Eagle Strategy and CMA-ES, respectively. Note that the good performance of RIBBO does not owe to the memorization of optimal solutions, as the search space is transformed randomly, resulting in variations in optimal solutions across different functions from the same distribution. It is because RIBBO is capable of using RTG tokens to identify algorithms and reinforce the performance on top of the behavior algorithms, which will be clearly shown later. We can also observe that RIBBO performs extremely well in the early stage, which draws the advantage from the HRR strategy, i.e., employing $0$ as the immediate RTG to generate the optimal potential solutions. 

RIBBO does not perform well on the SVM problem, which may be due to the problem's low-dimensional nature (only three parameters) and its relative simplicity for optimization. Behavior algorithms can achieve good performance easily, while the complexity of RIBBO's training and inference processes could instead result in the performance degradation. For the rover problem where GP-EI performs the best, we collect less data from GP-EI than other behavior algorithms due to the high time cost. This may limit RIBBO's capacity to leverage the high-quality data from GP-EI, given its small proportion relative to the data collected from other behavior algorithms. Despite this, RIBBO is still the runner-up, significantly surpassing the other behavior algorithms. 

Compared with BC and BC Filter, RIBBO performs consistently better except on the SVM problem. BC tends to imitate the average behavior of various algorithms, and its poor performance is due to the aggregation of behavior algorithms with inferior performance. BC Filter is generally better than BC, because the data from the two underperforming behavior algorithms, i.e., Random Search and Shuffled Grid Search, are manually excluded for the training of BC Filter. As introduced before, OptFormer requires to manually specify which algorithm to execute. We have specified the behavior algorithm Eagle Strategy in Figure~\ref{fig:main_exp}, which obtains good overall performance on these problems. It can be observed that OptFormer displays a close performance to Eagle Strategy, while RIBBO performs better. More results about the imitation capacity of OptFormer can be found in Appendix~\ref{appendix:optformer}.

\begin{figure*}[!t]
    \subfigure[Visualization]{\label{fig:visualization}\includegraphics[width=0.25\textwidth]{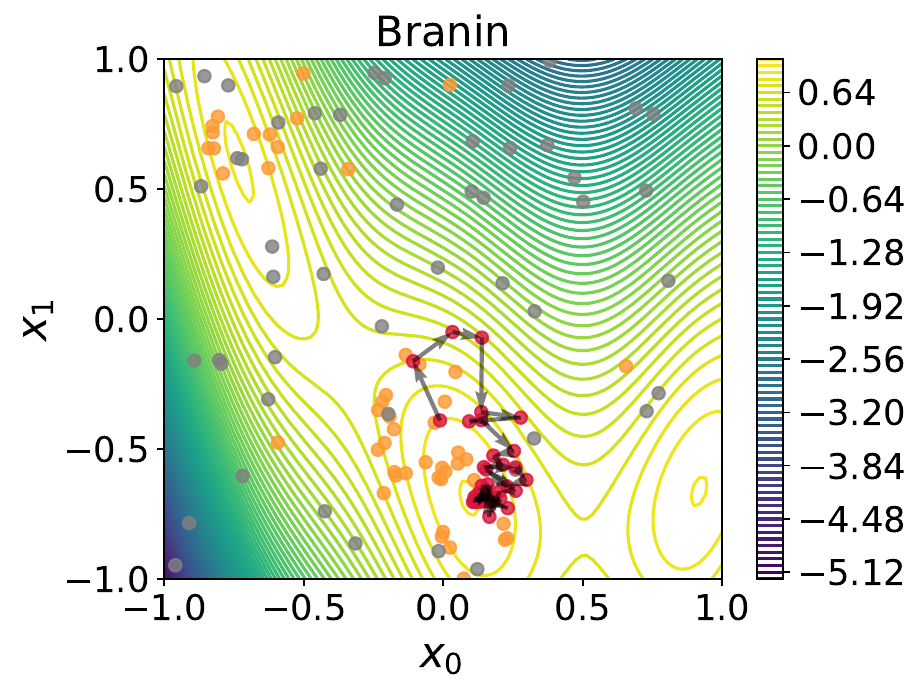}}
    \subfigure[Generalization]{\label{fig:generalization}\includegraphics[width=0.24\textwidth]{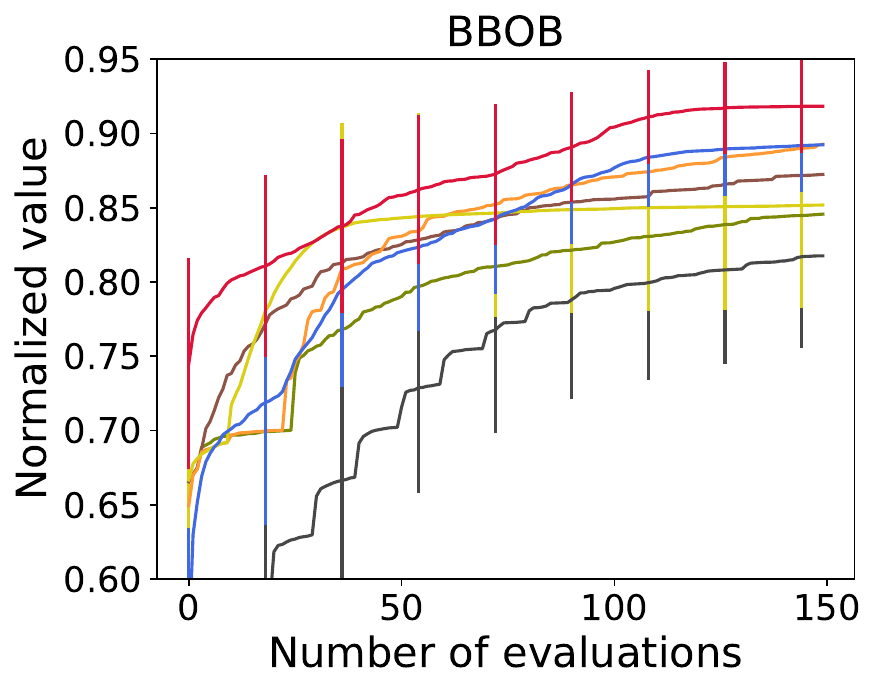}}
    \subfigure[Initial RTGs]{\label{fig:rtg-value}\includegraphics[width=0.231\textwidth]{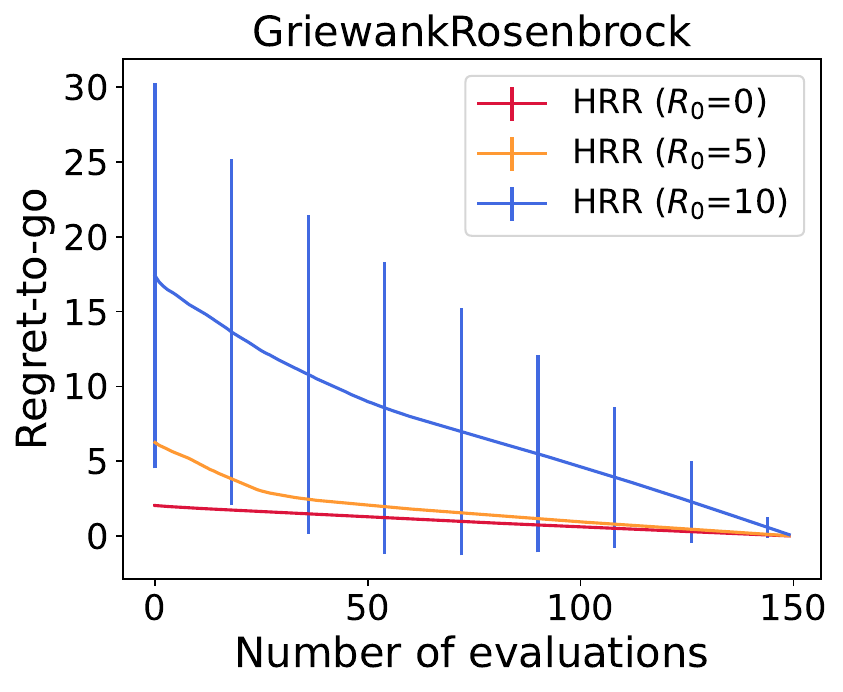}}
    \subfigure[RTG update strategy]{\label{fig:regret_strategy}\includegraphics[width=0.24\textwidth]{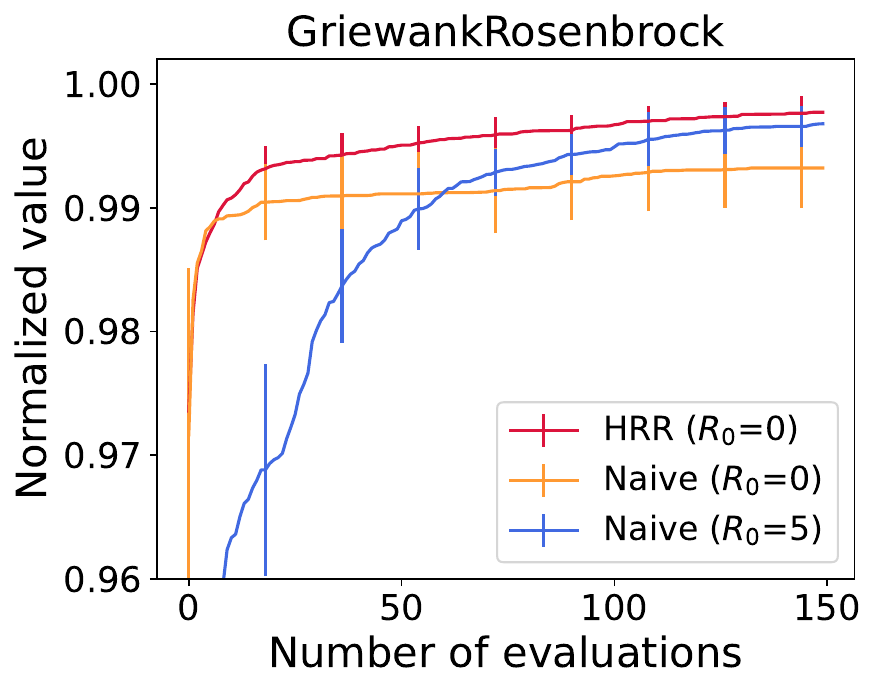}}
    \vspace{-0.7em}
    \caption{\textit{(a)~Visualization} of the contour lines of $2$D Branin function and sampling points of RIBBO (red), Eagle Strategy (orange), and Random Search (gray), where the arrows represent the optimization trajectory of RIBBO. \textit{(b)~Generalization} by training the model across all $24$ BBOB synthetic functions simultaneously. The results across functions with different output scaling are normalized to obtain the aggregate results. The legend shares with that of Figure~\ref{fig:main_exp}. \textit{(c)~Initial RTG} $R_0$'s influence on performance. \textit{(d)~RTG update strategy} comparison between HRR and the naive strategy with various initial RTG $R_0$. }\vspace{-0.9em}
\end{figure*}

\textbf{Why Does RIBBO Behave Well?} To better understand RIBBO, we train the model using only two behavior algorithms, Eagle Strategy and Random Search, which represent a good algorithm and an underperforming one, respectively. Figure~\ref{fig:visualization} visualizes the contour lines of the $2$D Branin function and the sampling points of RIBBO, Eagle Strategy, and Random Search, represented by red, orange, and gray points, respectively. The arrows are used to represent the optimization trajectory of RIBBO. Note that the two parameters of Branin have been scaled to $[-1, 1]$ for better visualization. It can be observed that RIBBO makes a prediction preferring Eagle Strategy over Random Search, indicating its capability to automatically identify the quality of training data. Additionally, RIBBO achieves the exploration and exploitation trade-off capability using its knowledge about the task obtained during training, thus generating superior solutions over the ones in the training dataset.

\textbf{Generalization.} We also conduct experiments to examine the generalization capabilities of RIBBO. For this purpose, we train the model on all $24$ BBOB synthetic functions simultaneously with the results shown in Figure~\ref{fig:generalization}. To aggregate results across functions with different output scaling, we normalize all the functions adhering to previous literature~\cite{DBLP:conf/nips/TurnerEMKLXG20,arango2021hpob,chen2022towards}. The results suggest that RIBBO demonstrates strong generalizing to a variety of functions with different properties. Further experiments are conducted to examine the cross-distribution generalization to unseen function distributions. The model is trained on $4$ of the $5$ chosen synthetic functions and tested with the remaining one. Note that each function (i.e., Greiwank Rosenbrock, Lunacek, Rastrigin, Rosenbrock, and Sharp Ridge) here actually represents a distribution of functions with similar properties, and a set of functions is sampled from each distribution as introduced before. The results suggest that RIBBO has a strong generalizing ability to unseen function distributions. Due to the space limitation, the results and more details on cross-distribution generalization are deferred to Appendix~\ref{appendix:generalization}. The good generalization of RIBBO can be attributed to the paradigm of learning the entire algorithm, which can acquire general knowledge, such as exploration and exploitation trade-off from data, as observed in~\cite{chen2017learning}. In contrast, such generalization may be limited if we learn surrogate models from data, because the function landscape inherent to surrogate models will contain only the knowledge of similar functions.


\subsection{Ablation Studies}
\label{sec:rtg-analysis}

RIBBO augments the histories with RTG tokens, facilitating distinguishing algorithms and automatically generating algorithms with user-specified performance. Next, we will verify the effectiveness of RTG conditioning and HRR strategy.

\textbf{Influence of Initial RTG Token $R_0$.} By incorporating RTG tokens, RIBBO is able to attend to RTGs and generate optimization trajectories based on the specified initial RTG token $R_0$. To validate this, we examine the performance of RIBBO with different values of $R_0$, and the results are presented in Figure~\ref{fig:rtg-value}. Here, the RTG values, instead of normalized objective values, are used as the $y$-axis. We can observe that the cumulative regrets of the generated query sequence do correlate with the specified RTG, indicating that RIBBO establishes the connection between regret and generation. We also conduct experiments to explore the effect of varying the immediate RTG $R_t$, and it is observed that setting to a value larger than $0$ will decrease the performance and converge to a worse value. Please see Appendix~\ref{appendix:rtg-diss}. 

\textbf{Effectiveness of HRR.} A key point of the inference procedure is how to update the value of RTG tokens at each iteration. To assess the effectiveness of the proposed strategy HRR outlined in Eq.~(\refeq{eq-hrr}), we compare it with the naive strategy, that sets an initial RTG token $R_0$ and decreases it by the one-step regret after each iteration. The results are shown in Figure~\ref{fig:regret_strategy}. The naive strategy displays distinct behaviors depending on the initial setting of $R_0$. Specifically, when $R_0=0$, i.e., the lower bound of regret, the model performs well initially. However, as the optimization progresses, the RTG tokens gradually decrease to negative values, leading to poor performance since negative RTGs are out-of-distribution values. Using $R_0=5$ compromises the initial performance, as the model may not select the most aggressive solutions with a high $R_0$. However, a higher initial $R_0$ yields better convergence value since it prevents out-of-distribution RTGs in later stage. The proposed HRR strategy consistently outperforms across the whole optimization stage, because setting the immediate RTG to $0$ encourages the model to make the most advantageous decisions at every iteration, while hindsight relabeling of previous RTG tokens, as specified in Eq.~(\refeq{eq-hrr}), ensures that these values remain meaningful and feasible.


\textbf{Further Studies.} We also study the effects of the method to aggregate $(\bm x_i, y_i, R_i)$ tokens, the normalization method for $y$, the model size, and the sampled subsequence length $\tau$. Please see Appendix~\ref{appendix:ablation}. More visualizations illustrating the effects of random transformations on the search space are detailed in Appendix~\ref{appendix:branin}.

\section{Conclusion}
This paper proposes RIBBO, which employs a transformer architecture to learn a reinforced BBO algorithm from offline datasets in an E2E fashion. By incorporating RTG tokens into the optimization histories, RIBBO can automatically generate optimization trajectories satisfying the user-desired regret. Comprehensive experiments on BBOB synthetic functions, HPO and robot control problems show the versatility of RIBBO. This work is a preliminary attempt towards universal BBO, and we hope it can encourage more explorations in this direction. For example, we only consider the model training over continuous search space with the same dimensionality, and it would be interesting to explore heteroscedastic search space with different types of variables. A mathematical theoretical analysis of the in-context learning capabilities based on RTG tokens is also of interest.

\bibliography{iclr2025_conference}

\begin{thebibliography}{70}
\providecommand{\natexlab}[1]{#1}
\providecommand{\url}[1]{\texttt{#1}}
\expandafter\ifx\csname urlstyle\endcsname\relax
  \providecommand{\doi}[1]{doi: #1}\else
  \providecommand{\doi}{doi: \begingroup \urlstyle{rm}\Url}\fi

\bibitem[Adriaensen et~al.(2022)Adriaensen, Biedenkapp, Shala, Awad, Eimer, Lindauer, and Hutter]{DBLP:journals/jair/AdriaensenBSAEL22}
Steven Adriaensen, Andr{\'{e}} Biedenkapp, Gresa Shala, Noor~H. Awad, Theresa Eimer, Marius Lindauer, and Frank Hutter.
\newblock Automated dynamic algorithm configuration.
\newblock \emph{Journal of Artificial Intelligence Research}, 75:\penalty0 1633--1699, 2022.

\bibitem[Alarie et~al.(2021)Alarie, Audet, Gheribi, Kokkolaras, and {Le Digabel}]{alarie2021two}
Stéphane Alarie, Charles Audet, Aïmen~E. Gheribi, Michael Kokkolaras, and Sébastien {Le Digabel}.
\newblock Two decades of blackbox optimization applications.
\newblock \emph{EURO Journal on Computational Optimization}, 9:\penalty0 100011, 2021.

\bibitem[Arango et~al.(2021)Arango, Jomaa, Wistuba, and Grabocka]{arango2021hpob}
Sebastian~Pineda Arango, Hadi~S Jomaa, Martin Wistuba, and Josif Grabocka.
\newblock {HPO-B:} {A} large-scale reproducible benchmark for black-box {HPO} based on {O}pen{ML}.
\newblock \emph{arXiv preprint arXiv:2106.06257}, 2021.

\bibitem[Astudillo \& Frazier(2021)Astudillo and Frazier]{10.5555/3522802.3522804}
Raul Astudillo and Peter~I. Frazier.
\newblock Thinking inside the box: {A} tutorial on grey-box {B}ayesian optimization.
\newblock In \emph{Proceedings of the 51st Winter Simulation Conference (WSC'21)}, pp.\  1--15, Phoenix, {AZ}, 2021.

\bibitem[Audet \& Hare(2017)Audet and Hare]{audet2017derivative}
Charles Audet and Warren Hare.
\newblock \emph{Derivative-Free and Blackbox Optimization}.
\newblock Springer, 2017.

\bibitem[Back(1996)]{back:96}
Thomas Back.
\newblock \emph{Evolutionary Algorithms in Theory and Practice: Evolution Strategies, Evolutionary Programming, Genetic Algorithms}.
\newblock Oxford University Press, 1996.

\bibitem[Bai et~al.(2023)Bai, Li, Shen, Zhang, Zhang, and Cui]{bai2023transfer}
Tianyi Bai, Yang Li, Yu~Shen, Xinyi Zhang, Wentao Zhang, and Bin Cui.
\newblock Transfer learning for {B}ayesian optimization: A survey.
\newblock \emph{arXiv preprint arXiv:2302.05927}, 2023.

\bibitem[Bain \& Sammut(1995)Bain and Sammut]{bain1995framework}
Michael Bain and Claude Sammut.
\newblock A framework for behavioural cloning.
\newblock \emph{Machine Intelligence}, 15:\penalty0 103--129, 1995.

\bibitem[Balandat et~al.(2020)Balandat, Karrer, Jiang, Daulton, Letham, Wilson, and Bakshy]{balandat2020botorch}
Maximilian Balandat, Brian Karrer, Daniel~R. Jiang, Samuel Daulton, Benjamin Letham, Andrew~Gordon Wilson, and Eytan Bakshy.
\newblock {Bo{T}orch: {A} framework for efficient {M}onte-{C}arlo {B}ayesian optimization}.
\newblock In \emph{Advances in Neural Information Processing Systems 33 (NeurIPS'20)}, pp.\  10113--10124, Virtual, 2020.

\bibitem[Biedenkapp et~al.(2020)Biedenkapp, Bozkurt, Eimer, Hutter, and Lindauer]{DBLP:conf/ecai/BiedenkappBEHL20}
Andr{\'{e}} Biedenkapp, H.~Furkan Bozkurt, Theresa Eimer, Frank Hutter, and Marius Lindauer.
\newblock Dynamic algorithm configuration: Foundation of a new meta-algorithmic framework.
\newblock In \emph{Proceedings of the 24th European Conference on Artificial Intelligence (ECAI'20)}, pp.\  427--434, Santiago, Spain, 2020.

\bibitem[Brown et~al.(2020)Brown, Mann, Ryder, Subbiah, Kaplan, Dhariwal, Neelakantan, Shyam, Sastry, Askell, Agarwal, Herbert{-}Voss, Krueger, Henighan, Child, Ramesh, Ziegler, Wu, Winter, Hesse, Chen, Sigler, Litwin, Gray, Chess, Clark, Berner, McCandlish, Radford, Sutskever, and Amodei]{NEURIPS2020_1457c0d6}
Tom~B. Brown, Benjamin Mann, Nick Ryder, Melanie Subbiah, Jared Kaplan, Prafulla Dhariwal, Arvind Neelakantan, Pranav Shyam, Girish Sastry, Amanda Askell, Sandhini Agarwal, Ariel Herbert{-}Voss, Gretchen Krueger, Tom Henighan, Rewon Child, Aditya Ramesh, Daniel~M. Ziegler, Jeffrey Wu, Clemens Winter, Christopher Hesse, Mark Chen, Eric Sigler, Mateusz Litwin, Scott Gray, Benjamin Chess, Jack Clark, Christopher Berner, Sam McCandlish, Alec Radford, Ilya Sutskever, and Dario Amodei.
\newblock Language models are few-shot learners.
\newblock In \emph{Advances in Neural Information Processing Systems 33 (NeurIPS'20)}, pp.\  1877--1901, Virtual, 2020.

\bibitem[Calandra et~al.(2016)Calandra, Seyfarth, Peters, and Deisenroth]{calandra2016bayesian}
Roberto Calandra, Andr{\'{e}} Seyfarth, Jan Peters, and Marc~Peter Deisenroth.
\newblock Bayesian optimization for learning gaits under uncertainty - {A}n experimental comparison on a dynamic bipedal walker.
\newblock \emph{Annals of Mathematics and Artificial Intelligence}, 76\penalty0 (1-2):\penalty0 5--23, 2016.

\bibitem[Chatzilygeroudis et~al.(2019)Chatzilygeroudis, Vassiliades, Stulp, Calinon, and Mouret]{8944013}
Konstantinos Chatzilygeroudis, Vassilis Vassiliades, Freek Stulp, Sylvain Calinon, and Jean-Baptiste Mouret.
\newblock A survey on policy search algorithms for learning robot controllers in a handful of trials.
\newblock \emph{IEEE Transactions on Robotics}, 36\penalty0 (2):\penalty0 328--347, 2019.

\bibitem[Chen et~al.(2021)Chen, Lu, Rajeswaran, Lee, Grover, Laskin, Abbeel, Srinivas, and Mordatch]{chen2021decision}
Lili Chen, Kevin Lu, Aravind Rajeswaran, Kimin Lee, Aditya Grover, Michael Laskin, Pieter Abbeel, Aravind Srinivas, and Igor Mordatch.
\newblock Decision transformer: Reinforcement learning via sequence modeling.
\newblock In \emph{Advances in Neural Information Processing Systems 34 (NeurIPS'21)}, pp.\  15084--15097, Virtual, 2021.

\bibitem[Chen et~al.(2017)Chen, Hoffman, Colmenarejo, Denil, Lillicrap, Botvinick, and Freitas]{chen2017learning}
Yutian Chen, Matthew~W Hoffman, Sergio~G{\'o}mez Colmenarejo, Misha Denil, Timothy~P Lillicrap, Matt Botvinick, and Nando Freitas.
\newblock Learning to learn without gradient descent by gradient descent.
\newblock In \emph{Proceedings of the 34th International Conference on Machine Learning (ICML'17)}, pp.\  748--756, Sydney, Australia, 2017.

\bibitem[Chen et~al.(2022)Chen, Song, Lee, Wang, Zhang, Dohan, Kawakami, Kochanski, Doucet, Ranzato, Perel, and de~Freitas]{chen2022towards}
Yutian Chen, Xingyou Song, Chansoo Lee, Zi~Wang, Richard Zhang, David Dohan, Kazuya Kawakami, Greg Kochanski, Arnaud Doucet, Marc'Aurelio Ranzato, Sagi Perel, and Nando de~Freitas.
\newblock Towards learning universal hyperparameter optimizers with transformers.
\newblock In \emph{Advances in Neural Information Processing Systems 35 (NeurIPS'22)}, pp.\  32053--32068, New Orleans, LA, 2022.

\bibitem[Elhara et~al.(2019)Elhara, Varelas, Nguyen, Tusar, Brockhoff, Hansen, and Auger]{ElHara2019COCOTL}
Ouassim Elhara, Konstantinos Varelas, Duc Nguyen, Tea Tusar, Dimo Brockhoff, Nikolaus Hansen, and Anne Auger.
\newblock {COCO}: {T}he large scale black-box optimization benchmarking ({BBOB}-largescale) test suite.
\newblock \emph{arXiv preprint arXiv:1903.06396}, 2019.

\bibitem[Eriksson et~al.(2019)Eriksson, Pearce, Gardner, Turner, and Poloczek]{eriksson2019scalable}
David Eriksson, Michael Pearce, Jacob~R. Gardner, Ryan Turner, and Matthias Poloczek.
\newblock Scalable global optimization via local {B}ayesian optimization.
\newblock In \emph{Advances in Neural Information Processing Systems 32 (NeurIPS'19)}, pp.\  5497--5508, Vancouver, Canada, 2019.

\bibitem[Feurer et~al.(2015)Feurer, Springenberg, and Hutter]{Feurer_Springenberg_Hutter_2015}
Matthias Feurer, Jost Springenberg, and Frank Hutter.
\newblock Initializing {B}ayesian hyperparameter optimization via meta-learning.
\newblock In \emph{Proceedings of the 29th AAAI Conference on Artificial Intelligence (AAAI'15)}, pp.\  1128--1135, Austin, TX, 2015.

\bibitem[Feurer et~al.(2021)Feurer, Van~Rijn, Kadra, Gijsbers, Mallik, Ravi, M{\"u}ller, Vanschoren, and Hutter]{OpenMLPython2019}
Matthias Feurer, Jan~N Van~Rijn, Arlind Kadra, Pieter Gijsbers, Neeratyoy Mallik, Sahithya Ravi, Andreas M{\"u}ller, Joaquin Vanschoren, and Frank Hutter.
\newblock Open{ML}-{P}ython: {A}n extensible {P}ython {API} for {O}pen{ML}.
\newblock \emph{The Journal of Machine Learning Research}, 22\penalty0 (1):\penalty0 4573--4577, 2021.

\bibitem[Finn et~al.(2017)Finn, Abbeel, and Levine]{finn2017model}
Chelsea Finn, Pieter Abbeel, and Sergey Levine.
\newblock Model-agnostic meta-learning for fast adaptation of deep networks.
\newblock In \emph{Proceedings of the 34th International Conference on Machine Learning (ICML'17)}, pp.\  1126--1135, Sydney, Australia, 2017.

\bibitem[Frazier(2018)]{bosurvey2}
Peter~I Frazier.
\newblock A tutorial on {B}ayesian optimization.
\newblock \emph{arXiv preprint arXiv:1807.02811}, 2018.

\bibitem[Frazier \& Wang(2016)Frazier and Wang]{frazier2015bayesian}
Peter~I Frazier and Jialei Wang.
\newblock \emph{Bayesian Optimization for Materials Design}.
\newblock Springer, 2016.

\bibitem[Garnelo et~al.(2018)Garnelo, Rosenbaum, Maddison, Ramalho, Saxton, Shanahan, Teh, Rezende, and Eslami]{garnelo2018conditional}
Marta Garnelo, Dan Rosenbaum, Christopher Maddison, Tiago Ramalho, David Saxton, Murray Shanahan, Yee~Whye Teh, Danilo Rezende, and SM~Ali Eslami.
\newblock Conditional neural processes.
\newblock In \emph{Proceedings of the 35th International Conference on Machine Learning (ICML'18)}, pp.\  1704--1713, Stockholm, Sweden, 2018.

\bibitem[G{\'o}mez-Bombarelli et~al.(2018)G{\'o}mez-Bombarelli, Duvenaud, Hern{\'a}ndez-Lobato, Aguilera-Iparraguirre, Hirzel, Adams, and Aspuru-Guzik]{gomez2018automatic}
R.~G{\'o}mez-Bombarelli, D.~K. Duvenaud, J.~M. Hern{\'a}ndez-Lobato, J.~Aguilera-Iparraguirre, T.D. Hirzel, R.~P. Adams, and A.~Aspuru-Guzik.
\newblock Automatic chemical design using a data-driven continuous representation of molecules.
\newblock \emph{ACS Central Science}, 4\penalty0 (2):\penalty0 268 -- 276, 2018.

\bibitem[Greenhill et~al.(2020)Greenhill, Rana, Gupta, Vellanki, and Venkatesh]{greenhill2020bayesian}
Stewart Greenhill, Santu Rana, Sunil Gupta, Pratibha Vellanki, and Svetha Venkatesh.
\newblock Bayesian optimization for adaptive experimental design: {A} review.
\newblock \emph{{IEEE} Access}, 8:\penalty0 13937--13948, 2020.

\bibitem[Guo et~al.(2023)Guo, Hu, Mei, Wang, Xiong, Savarese, and Bai]{guo2023transformers}
Tianyu Guo, Wei Hu, Song Mei, Huan Wang, Caiming Xiong, Silvio Savarese, and Yu~Bai.
\newblock How do transformers learn in-context beyond simple functions? {A} case study on learning with representations.
\newblock \emph{arXiv preprint arXiv:2310.10616}, 2023.

\bibitem[Hansen(2016)]{cmaes}
Nikolaus Hansen.
\newblock The {CMA} evolution strategy: {A} tutorial.
\newblock \emph{arXiv preprint arXiv:1604.00772}, 2016.

\bibitem[Hollmann et~al.(2023)Hollmann, M{\"u}ller, Eggensperger, and Hutter]{hollmann2023tabpfn}
Noah Hollmann, Samuel M{\"u}ller, Katharina Eggensperger, and Frank Hutter.
\newblock Tab{PFN}: {A} transformer that solves small tabular classification problems in a second.
\newblock In \emph{Proceedings of the 11th International Conference on Learning Representations (ICLR'23)}, Kigali, Rwanda, 2023.

\bibitem[Hospedales et~al.(2021)Hospedales, Antoniou, Micaelli, and Storkey]{hospedales2021meta}
Timothy Hospedales, Antreas Antoniou, Paul Micaelli, and Amos Storkey.
\newblock Meta-learning in neural networks: A survey.
\newblock \emph{IEEE Transactions on Pattern Analysis and Machine Intelligence}, 44\penalty0 (9):\penalty0 5149--5169, 2021.

\bibitem[Hsieh et~al.(2021)Hsieh, Hsieh, and Liu]{hsieh2021reinforced}
Bing{-}Jing Hsieh, Ping{-}Chun Hsieh, and Xi~Liu.
\newblock Reinforced few-shot acquisition function learning for {B}ayesian optimization.
\newblock In \emph{Advances in Neural Information Processing Systems 34 (NeurIPS'21)}, pp.\  7718--7731, Virtual, 2021.

\bibitem[Jones et~al.(1998)Jones, Schonlau, and Welch]{jones1998efficient}
Donald~R Jones, Matthias Schonlau, and William~J Welch.
\newblock Efficient global optimization of expensive black-box functions.
\newblock \emph{Journal of Global Optimization}, 13\penalty0 (4):\penalty0 455--492, 1998.

\bibitem[Kaplan et~al.(2020)Kaplan, McCandlish, Henighan, Brown, Chess, Child, Gray, Radford, Wu, and Amodei]{Kaplan2020ScalingLF}
Jared Kaplan, Sam McCandlish, Tom Henighan, Tom~B Brown, Benjamin Chess, Rewon Child, Scott Gray, Alec Radford, Jeffrey Wu, and Dario Amodei.
\newblock Scaling laws for neural language models.
\newblock \emph{arXiv preprint arXiv:2001.08361}, 2020.

\bibitem[Khan et~al.(2022)Khan, Naseer, Hayat, Zamir, Khan, and Shah]{khan2022transformers}
Salman Khan, Muzammal Naseer, Munawar Hayat, Syed~Waqas Zamir, Fahad~Shahbaz Khan, and Mubarak Shah.
\newblock Transformers in vision: A survey.
\newblock \emph{ACM Computing Surveys}, 54\penalty0 (10s):\penalty0 1--41, 2022.

\bibitem[Krishnamoorthy et~al.(2023)Krishnamoorthy, Mashkaria, and Grover]{Krishnamoorthy2022GenerativePF}
Siddarth Krishnamoorthy, Satvik Mashkaria, and Aditya Grover.
\newblock Generative pretraining for black-box optimization.
\newblock In \emph{Proceedings of the 40th International Conference on Machine Learning (ICML'23)}, pp.\  24173--24197, Honolulu, HI, 2023.

\bibitem[Lange et~al.(2023{\natexlab{a}})Lange, Schaul, Chen, Lu, Zahavy, Dalibard, and Flennerhag]{lange2023discoveringga}
Robert~Tjarko Lange, Tom Schaul, Yutian Chen, Chris Lu, Tom Zahavy, Valentin Dalibard, and Sebastian Flennerhag.
\newblock Discovering attention-based genetic algorithms via meta-black-box optimization.
\newblock In \emph{Proceedings of the 25th Conference on Genetic and Evolutionary Computation (GECCO'23)}, pp.\  929--937, Lisbon, Portugal, 2023{\natexlab{a}}.

\bibitem[Lange et~al.(2023{\natexlab{b}})Lange, Schaul, Chen, Zahavy, Dalibard, Lu, Singh, and Flennerhag]{lange2023discoveringes}
Robert~Tjarko Lange, Tom Schaul, Yutian Chen, Tom Zahavy, Valentin Dalibard, Chris Lu, Satinder Singh, and Sebastian Flennerhag.
\newblock Discovering evolution strategies via meta-black-box optimization.
\newblock In \emph{Proceedings of the 11th International Conference on Learning Representations (ICLR'23)}, Kigali, Rwanda, 2023{\natexlab{b}}.

\bibitem[Laskin et~al.(2023)Laskin, Wang, Oh, Parisotto, Spencer, Steigerwald, Strouse, Hansen, Filos, Brooks, maxime gazeau, Sahni, Singh, and Mnih]{laskin2023incontext}
Michael Laskin, Luyu Wang, Junhyuk Oh, Emilio Parisotto, Stephen Spencer, Richie Steigerwald, DJ~Strouse, Steven~Stenberg Hansen, Angelos Filos, Ethan Brooks, maxime gazeau, Himanshu Sahni, Satinder Singh, and Volodymyr Mnih.
\newblock In-context reinforcement learning with algorithm distillation.
\newblock In \emph{Proceedings of the 11th International Conference on Learning Representations (ICLR'23)}, Kigali, Rwanda, 2023.

\bibitem[Li et~al.(2022)Li, Shen, Jiang, Bai, Zhang, Zhang, and Cui]{li2022transfer}
Yang Li, Yu~Shen, Huaijun Jiang, Tianyi Bai, Wentao Zhang, Ce~Zhang, and Bin Cui.
\newblock Transfer learning based search space design for hyperparameter tuning.
\newblock In \emph{Proceedings of the 28th {ACM} {SIGKDD} Conference on Knowledge Discovery and Data Mining (KDD'22)}, pp.\  967--977, Washington, DC, 2022.

\bibitem[Li et~al.(2023)Li, Ildiz, Papailiopoulos, and Oymak]{li2023transformers}
Yingcong Li, Muhammed~Emrullah Ildiz, Dimitris Papailiopoulos, and Samet Oymak.
\newblock Transformers as algorithms: Generalization and stability in in-context learning.
\newblock In \emph{Proceedings of the 40th International Conference on Machine Learning (ICML'23)}, pp.\  19565--19594, Honolulu, HI, 2023.

\bibitem[Maraval et~al.(2023)Maraval, Zimmer, Grosnit, and Ammar]{maraval2023endtoend}
Alexandre~Max Maraval, Matthieu Zimmer, Antoine Grosnit, and Haitham~Bou Ammar.
\newblock End-to-end meta-{B}ayesian optimisation with transformer neural processes.
\newblock In \emph{Advances in Neural Information Processing Systems 36 (NeurIPS'23)}, New Orleans, LA, 2023.

\bibitem[Müller et~al.(2022)Müller, Hollmann, Arango, Grabocka, and Hutter]{nokey}
Samuel Müller, Noah Hollmann, Sebastian~Pineda Arango, Josif Grabocka, and Frank Hutter.
\newblock Transformers can do {B}ayesian inference.
\newblock In \emph{Proceedings of the 10th International Conference on Learning Representations (ICLR'22)}, Virtual, 2022.

\bibitem[Müller et~al.(2023)Müller, Feurer, Hollmann, and Hutter]{sam_at_icml23}
Samuel Müller, Matthias Feurer, Noah Hollmann, and Frank Hutter.
\newblock {PFN}s4{BO}: {I}n-context learning for {B}ayesian optimization.
\newblock In \emph{Proceedings of the 40th International Conference on Machine Learning (ICML'23)}, pp.\  25444--25470, Honolulu, HI, 2023.

\bibitem[Negoescu et~al.(2011)Negoescu, Frazier, and Powell]{Negoescu2011TheKA}
Diana~M. Negoescu, Peter~I. Frazier, and Warren~B. Powell.
\newblock The knowledge-gradient algorithm for sequencing experiments in drug discovery.
\newblock \emph{{INFORMS} Journal on Computing}, 23\penalty0 (3):\penalty0 346--363, 2011.

\bibitem[Nguyen \& Grover(2022)Nguyen and Grover]{Nguyen2022TransformerNP}
Tung Nguyen and Aditya Grover.
\newblock Transformer neural processes: Uncertainty-aware meta learning via sequence modeling.
\newblock In \emph{Proceedings of the 39th International Conference on Machine Learning (ICML'22)}, pp.\  16569--16594, Baltimore, MD, 2022.

\bibitem[Nguyen et~al.(2023)Nguyen, Agrawal, and Grover]{nguyen2023expt}
Tung Nguyen, Sudhanshu Agrawal, and Aditya Grover.
\newblock Ex{PT}: Synthetic pretraining for few-shot experimental design.
\newblock \emph{arXiv preprint arXiv:2310.19961}, 2023.

\bibitem[Perrone \& Shen(2019)Perrone and Shen]{perrone2019learning}
Valerio Perrone and Huibin Shen.
\newblock Learning search spaces for {B}ayesian optimization: Another view of hyperparameter transfer learning.
\newblock In \emph{Advances in Neural Information Processing Systems 32 (NeurIPS'19)}, pp.\  12751--12761, Vancouver, Canada, 2019.

\bibitem[Perrone et~al.(2018)Perrone, Jenatton, Seeger, and Archambeau]{NEURIPS2018_14c879f3}
Valerio Perrone, Rodolphe Jenatton, Matthias~W. Seeger, and C{\'{e}}dric Archambeau.
\newblock Scalable hyperparameter transfer learning.
\newblock In \emph{Advances in Neural Information Processing Systems 31 (NeurIPS'18)}, pp.\  6846--6856, Montreal, Canada, 2018.

\bibitem[Poloczek et~al.(2016)Poloczek, Wang, and Frazier]{poloczek2016warm}
Matthias Poloczek, Jialei Wang, and Peter~I Frazier.
\newblock Warm starting {B}ayesian optimization.
\newblock In \emph{Proceedings of the 46th Winter Simulation Conference (WSC'16)}, pp.\  770--781, Washington, DC, 2016.

\bibitem[Radford et~al.(2018)Radford, Narasimhan, Salimans, and Sutskever]{radford2018improving}
Alec Radford, Karthik Narasimhan, Tim Salimans, and Ilya Sutskever.
\newblock Improving language understanding by generative pre-training.
\newblock \emph{OpenAI Blog}, 2018.

\bibitem[Rasmussen \& Williams(2006)Rasmussen and Williams]{gpml}
C.~E. Rasmussen and C.~K.~I. Williams.
\newblock \emph{{G}aussian {P}rocesses for {M}achine {L}earning}.
\newblock The MIT Press, 2006.

\bibitem[Real et~al.(2019)Real, Aggarwal, Huang, and Le]{real2019regularized}
Esteban Real, Alok Aggarwal, Yanping Huang, and Quoc~V Le.
\newblock Regularized evolution for image classifier architecture search.
\newblock In \emph{Proceedings of the 33rd AAAI Conference on Artificial Intelligence (AAAI'19)}, pp.\  4780--4789, Honolulu, HI, 2019.

\bibitem[Shahriari et~al.(2016)Shahriari, Swersky, Wang, Adams, and de~Freitas]{bosurvey1}
Bobak Shahriari, Kevin Swersky, Ziyu Wang, Ryan~P. Adams, and Nando de~Freitas.
\newblock Taking the human out of the loop: {A} review of {B}ayesian optimization.
\newblock \emph{Proceedings of the IEEE}, 104\penalty0 (1):\penalty0 148--175, 2016.

\bibitem[Song et~al.(2022)Song, Perel, Lee, Kochanski, and Golovin]{oss_vizier}
Xingyou Song, Sagi Perel, Chansoo Lee, Greg Kochanski, and Daniel Golovin.
\newblock Open {S}ource {V}izier: Distributed infrastructure and {API} for reliable and flexible black-box optimization.
\newblock In \emph{Proceedings of the 1st International Conference on Automated Machine Learning (AutoML Conference'22)}, pp.\  1--17, Baltimore, MD, 2022.

\bibitem[Terayama et~al.(2021)Terayama, Sumita, Tamura, and Tsuda]{terayama2021black}
Kei Terayama, Masato Sumita, Ryo Tamura, and Koji Tsuda.
\newblock Black-box optimization for automated discovery.
\newblock \emph{Accounts of Chemical Research}, 54\penalty0 (6):\penalty0 1334--1346, 2021.

\bibitem[Turner et~al.(2020)Turner, Eriksson, McCourt, Kiili, Laaksonen, Xu, and Guyon]{DBLP:conf/nips/TurnerEMKLXG20}
Ryan Turner, David Eriksson, Michael McCourt, Juha Kiili, Eero Laaksonen, Zhen Xu, and Isabelle Guyon.
\newblock Bayesian optimization is superior to random search for machine learning hyperparameter tuning: Analysis of the black-box optimization challenge 2020.
\newblock In Hugo~Jair Escalante and Katja Hofmann (eds.), \emph{Advances in Neural Information Processing Systems 33 (NeurIPS'20) Competition and Demonstration Track}, pp.\  3--26, Virtual, 2020.

\bibitem[TV et~al.(2019)TV, Malhotra, Narwariya, Vig, and Shroff]{tv2019meta}
Vishnu TV, Pankaj Malhotra, Jyoti Narwariya, Lovekesh Vig, and Gautam Shroff.
\newblock Meta-learning for black-box optimization.
\newblock In \emph{Proceedings of European Conference on Machine Learning and Knowledge Discovery in Databases (ECML PKDD'19)}, pp.\  366--381, Würzburg, Germany, 2019.

\bibitem[Vaswani et~al.(2017)Vaswani, Shazeer, Parmar, Uszkoreit, Jones, Gomez, Kaiser, and Polosukhin]{vaswani2017attention}
Ashish Vaswani, Noam Shazeer, Niki Parmar, Jakob Uszkoreit, Llion Jones, Aidan~N. Gomez, Lukasz Kaiser, and Illia Polosukhin.
\newblock Attention is all you need.
\newblock In \emph{Advances in Neural Information Processing Systems 30 (NIPS'17)}, pp.\  5998--6008, Long Beach, Canada, 2017.

\bibitem[Vilalta \& Drissi(2002)Vilalta and Drissi]{vilalta2002perspective}
Ricardo Vilalta and Youssef Drissi.
\newblock A perspective view and survey of meta-learning.
\newblock \emph{Artificial Intelligence Review}, 18\penalty0 (2):\penalty0 77--95, 2002.

\bibitem[Volpp et~al.(2020)Volpp, Fr{\"{o}}hlich, Fischer, Doerr, Falkner, Hutter, and Daniel]{Volpp2020Meta-Learning}
Michael Volpp, Lukas~P. Fr{\"{o}}hlich, Kirsten Fischer, Andreas Doerr, Stefan Falkner, Frank Hutter, and Christian Daniel.
\newblock Meta-learning acquisition functions for transfer learning in {B}ayesian optimization.
\newblock In \emph{Proceedings of the 8th International Conference on Learning Representations (ICLR'20)}, Addis Ababa, Ethiopia, 2020.

\bibitem[Wang et~al.(2018)Wang, Gehring, Kohli, and Jegelka]{wang2018batched}
Zi~Wang, Clement Gehring, Pushmeet Kohli, and Stefanie Jegelka.
\newblock Batched large-scale {B}ayesian optimization in high-dimensional spaces.
\newblock In \emph{Proceedings of the 21st International Conference on Artificial Intelligence and Statistics (AISTATS'18)}, pp.\  745--754, Playa Blanca, Spain, 2018.

\bibitem[Wang et~al.(2021)Wang, Dahl, Swersky, Lee, Nado, Gilmer, Snoek, and Ghahramani]{wang2021pre}
Zi~Wang, George~E Dahl, Kevin Swersky, Chansoo Lee, Zachary Nado, Justin Gilmer, Jasper Snoek, and Zoubin Ghahramani.
\newblock Pre-trained {G}aussian processes for {B}ayesian optimization.
\newblock \emph{arXiv preprint arXiv:2109.08215}, 2021.

\bibitem[Wen et~al.(2022)Wen, Zhou, Zhang, Chen, Ma, Yan, and Sun]{wen2022transformers}
Qingsong Wen, Tian Zhou, Chaoli Zhang, Weiqi Chen, Ziqing Ma, Junchi Yan, and Liang Sun.
\newblock Transformers in time series: A survey.
\newblock \emph{arXiv preprint arXiv:2202.07125}, 2022.

\bibitem[Wilson et~al.(2018)Wilson, Hutter, and Deisenroth]{wilson2018maximizing}
James~T. Wilson, Frank Hutter, and Marc~Peter Deisenroth.
\newblock Maximizing acquisition functions for {B}ayesian optimization.
\newblock In \emph{Advances in Neural Information Processing Systems 31 (NeurIPS'18)}, pp.\  9906--9917, Montr{\'{e}}al, Canada, 2018.

\bibitem[Wistuba \& Grabocka(2021)Wistuba and Grabocka]{wistuba2021fewshot}
Martin Wistuba and Josif Grabocka.
\newblock Few-shot {B}ayesian optimization with deep kernel surrogates.
\newblock In \emph{Proceedings of the 9th International Conference on Learning Representations (ICLR'21)}, Virtual, 2021.

\bibitem[Wolf et~al.(2020)Wolf, Debut, Sanh, Chaumond, Delangue, Moi, Cistac, Rault, Louf, Funtowicz, Davison, Shleifer, von Platen, Ma, Jernite, Plu, Xu, Scao, Gugger, Drame, Lhoest, and Rush]{wolf2020transformers}
Thomas Wolf, Lysandre Debut, Victor Sanh, Julien Chaumond, Clement Delangue, Anthony Moi, Pierric Cistac, Tim Rault, R{\'{e}}mi Louf, Morgan Funtowicz, Joe Davison, Sam Shleifer, Patrick von Platen, Clara Ma, Yacine Jernite, Julien Plu, Canwen Xu, Teven~Le Scao, Sylvain Gugger, Mariama Drame, Quentin Lhoest, and Alexander~M. Rush.
\newblock Transformers: {S}tate-of-the-art natural language processing.
\newblock In \emph{Proceedings of the 25th Conference on Empirical Methods in Natural Language Processing: System Demonstrations (EMNLP'20)}, pp.\  38--45, Virtual, 2020.

\bibitem[Yang \& Deb(2010)Yang and Deb]{yang2010eagle}
Xin-She Yang and Suash Deb.
\newblock Eagle strategy using {L}{\'e}vy walk and firefly algorithms for stochastic optimization.
\newblock In \emph{Proceedings of the 4th Nature Inspired Cooperative Strategies for Optimization (NICSO'10)}, pp.\  101--111, Granada, Spain, 2010.

\bibitem[Zhao et~al.(2024)Zhao, Liu, Yan, Duan, Yang, and Shi]{DBLP:journals/corr/abs-2405-03419}
Qi~Zhao, Tengfei Liu, Bai Yan, Qiqi Duan, Jian Yang, and Yuhui Shi.
\newblock Automated metaheuristic algorithm design with autoregressive learning.
\newblock \emph{arXiv preprint arXiv:2405.03419}, 2024.

\bibitem[Zhou et~al.(2023)Zhou, Wu, Song, Cao, and Zhang]{omni-vrp}
Jianan Zhou, Yaoxin Wu, Wen Song, Zhiguang Cao, and Jie Zhang.
\newblock Towards omni-generalizable neural methods for vehicle routing problems.
\newblock In \emph{Proceedings of the 40th International Conference on Machine Learning (ICML'23)}, pp.\  42769--42789, Honolulu, HI, 2023.

\bibitem[Zhou et~al.(2019)Zhou, Yu, and Qian]{zhou2019evolutionary}
Zhi-Hua Zhou, Yang Yu, and Chao Qian.
\newblock \emph{Evolutionary Learning: Advances in Theories and Algorithms}.
\newblock Springer, 2019.

\end{thebibliography}
\bibliographystyle{iclr2025_conference}

\appendix

\section{Model Details}
\label{appendix:model-details}

We employ the commonly used GPT architecture~\cite{radford2018improving} and the hyper-parameters are maintained consistently across the problems. Details can be found in Table~\ref{tab:hp}. For BC, BC Filter and OptFormer, the hyper-parameters are set as same as those of our model. The training takes about $20$ hours on $1$ GPU (Nvidia RTX $4090$).

\begin{table*}[ht]
\caption{List of hyper-parameter settings in RIBBO.}
\label{tab:hp}
\vskip 0.03in
\begin{center}
\begin{normalsize}
\begin{tabular}{lc}
\toprule
\multicolumn{2}{c}{RIBBO} \\
\midrule
Embedding dimension & 256 \\
Number of self-attention layers & 12 \\
Number of self-attention heads & 8 \\
Point-wise feed-forward dimension & 1024 \\
Dropout rate & 0.1 \\ 
Batch size & 64 \\
Learning rate & 0.0002 \\
Learning rate decay & 0.01 \\
Optimizer & Adam \\
Optimizer scheduler & Linear warm up and cosine annealing \\
Number of training steps & 500,000 \\
Length of subsequence $\tau$ & 50 \\
\bottomrule
\end{tabular}
\end{normalsize}
\end{center}
\vskip -0.1in
\end{table*}

\section{Details of Experimental Setup}
\label{appendix:experimental-setup}

\subsection{Benchmarks}
\label{appendix:benchmarks}

\begin{itemize}
    \item \textbf{BBOB}~\cite{ElHara2019COCOTL} is a widely used synthetic BBO benchmark, consisting of $24$ synthetic functions in the continuous domain. This benchmark makes a series of transformations in the search space, such as linear transformations (e.g., translation, rotation, scaling) and non-linear transformations (e.g., Tosz, Tasy), to create a distribution of functions while retaining similar properties. According to the properties of functions, these synthetic functions can be divided into $5$ categories, i.e., (1) separable functions, (2) moderately conditioned functions, (3) ill-conditioned and unimodal functions, (4) multi-modal functions with adequate global structure, and (5) multi-modal functions with weak global structures. We select one function from each category to evaluate our algorithm, Rastrigin from (1), Rosenbrock Rotated from (2), Sharp Ridge from (3), Greiwank Rosenbrock from (4), and Lunacek from (5). We use the BBOB benchmark implementation in Open Source Vizier\footnote{\url{https://github.com/google/vizier}}, and the dimension is set to $10$ for all functions. 

    \item \textbf{SVM and XGBoost.} HPO-B~\cite{arango2021hpob} is the most commonly used HPO benchmark and is grouped by search space id. Each search space id corresponds to a machine learning model, e.g., SVM or XGBoost. Each such search space has multiple associated dataset id, which is a particular HPO problem, i.e., optimizing the performance of the corresponding model on a dataset. For the continuous domain, it fits an XGBoost model as the objective function for each HPO problem. These datasets for each search space id are divided into training and test datasets. We examine our method on two selected search space id, i.e., $5527$ and $6767$, which are two representative HPO problems tuning SVM and XGBoost, respectively. SVM has $3$ parameters, while XGBoost has $18$ parameters, which is the most in HPO-B. We use the official implementations\footnote{\url{https://github.com/releaunifreiburg/HPO-B}}.
    

    \item \textbf{Rover Trajectory Planning}~\cite{eriksson2019scalable,wang2018batched} is a trajectory optimization task designed to emulate a rover navigation task. The trajectory is determined by fitting a B-spline to $30$ points in a $2$D plane, resulting in a total of $60$ parameters to optimize. Given $\bm x\in [0, 1]^{60}$, the objective function is $f(\bm x) = c(\bm x) + \lambda (\Vert \bm x_{0,1}-\bm s\Vert_1 + \Vert \bm x_{58,59}-\bm g\Vert_1) + b$, where $c(\bm x)$ is the cost of the given trajectory, $\bm x_{0,1}$ denotes the first and second dimensions of $\bm x$ (similarly for $\bm x_{58, 59}$), $\bm s$ and $\bm g$ are $2$D points that specify the starting and goal positions in the plane, $\lambda$ and $b$ are parameters to define the problem. This problem is non-smooth, discontinuous, and concave over the first two and last two dimensions. To construct the distribution of functions, we applied translations in $[-0.1, 0.1]^d$ and scalings in $[0.9, 1.1]$, similar to previous works~\cite{ElHara2019COCOTL,Volpp2020Meta-Learning}. The training and test tasks are randomly sampled from the distribution. We use the standard implementation for the rover problem\footnote{\url{https://github.com/uber-research/TuRBO}}. 
\end{itemize}

\subsection{Bahavior Algorithms}
\label{appendix:behavior-algo}

The datasets are generated with several representatives of heuristic search, EA, and BO as behavior algorithms. We use the implementation in Open Source Vizier~\cite{oss_vizier} for Random Search, Shuffled Grid Search, Eagle Strategy, CMA-ES. For Hill Climbing and Regularized Evolution, we provide a simple re-implementation. We use the implementation in BoTorch~\cite{balandat2020botorch} for GP-EI. The details of these behavior algorithms are summarized as follows.

\begin{itemize}
    \item Random Search selects a point uniformly at random from the domain at each iteration. 
    \item Shuffled Grid Search discretizes the ranges of real parameters into $100$ equidistant points and selects a random point from the grid without replacement at each iteration.
    \item Hill Climbing. At each iteration $t$, the current best solution $x_{best}$ is mutated (using the same mutation operation as Regularized Evolution) to generate $x_{next}$, which is then evaluated. If $f(x_{next}) > f(x_{best})$, $x_{best}$ is updated to $x_{next}$. 
    \item Regularized Evolution~\cite{real2019regularized} is an evolutionary algorithm with tournament selection and age-based replacement. We use a population size of $25$ and a tournament size of $5$. At each iteration, a tournament subset is randomly selected from the current population, and the solution with the maximum value is mutated. The mutation operation uniformly selects one of the parameters and mutates it to a random value within the domain.
    \item Eagle Strategy~\cite{yang2010eagle} without the Levy random walk, aka Firefly Algorithm, maintains a population of fireflies. Each firefly emits light whose intensity corresponds to the objective value. At each iteration, for each firefly, a weight is calculated to chase after a brighter firefly and actively move away from darker ones in its vicinity. The position is updated based on the calculated weight. 
    \item CMA-ES~\cite{cmaes} is a popular evolutionary algorithm. At each iteration, candidate solutions are sampled from a multivariate normal distribution and evaluated, then the mean and covariance matrix are updated.
    \item GP-EI employs GP as the surrogate model and EI~\cite{jones1998efficient} as the acquisition function for BO. 
\end{itemize}

RIBBO is trained exclusively from offline datasets derived from various source tasks, which are sampled from the task distribution. Each dataset consists of several optimization histories, generated by running a behavior algorithm on a task. No additional assumptions are imposed to the data collection process or the behavior algorithms, thereby permitting the use of any behavior algorithm.

Nevertheless, we can choose appropriate behavior algorithms to help the training. The model is trained on the datasets collected by the behavior algorithms, and the characteristics of these datasets can influence the model's efficacy. It is advisable to employ well-performing and diverse behavior algorithms to create these datasets. If the datasets are from suboptimal behavior algorithms predominantly, there might be a decline in performance, even with the incorporation of RTG tokens. If the datasets are predominantly produced by behavior algorithms with homogeneous properties, the model may only be adept at addressing specific problems characterized by those properties. Conversely, if the behavior algorithms are diverse, the model can learn the strengths from various algorithms.

\subsection{Data}
\label{appendix:data}

For BBOB and rover problem, a set of tasks is sampled from the task distribution and the above behavior algorithms are used to collect data. Specifically, for the BBOB suite, a total of $200$ functions are sampled, and each behavior algorithm is executed $500$ times, except for GP-EI, which is limited to $100$ function samples due to its high time cost. For the rover problem, $300$ functions are sampled with each being run $500$ times, whereas for GP-EI, a smaller number of $50$ functions is selected, each being run $500$ times. For testing, we randomly sample $10$ functions for each problem and run each algorithm multiple times to report the average performance and standard deviation. 

For HPO-B, we use the ``v3'' meta-training/test splits provided by the authors, which consist of $51$ training and $6$ test tasks for SVM, and similarly, $52$ training and $6$ test tasks for XGBoost. All behavior algorithms employ $500$ random seeds to collect the training datasets. 

\section{Re-Implementation of OptFormer}
\label{appendix:optformer}

OptFormer~\cite{chen2022towards} is a general optimization framework based on transformer~\cite{vaswani2017attention} and provides an interface to learn policy or function prediction. When provided with textual representations of historical data, OptFormer can determine the next query point $\bm x_t$, acting as a policy. Additionally, if the context incorporate a possible query point $\bm x_t$, the framework is capable of predicting the corresponding $y_t$, thus serving as a prediction model. We focus on the aspect of policy learning, due to its greater relevance to our work. 

The original implementation is built upon a text-based transformer and uses private datasets for training. In this paper, we have re-implemented a simplified version of OptFormer, where we omit the textual tokenization process and only retain the algorithm type within the metadata. Numerical inputs are fed into the model and we use different initial states $\bm x_0$ and $y_0$ to distinguish among algorithms. The initial states are derived by indexing the algorithm type in an embedding layer, thereby enabling the identification of distinct behavior algorithms within OptFormer. The hyper-parameters are set as same as our method. 

\begin{figure*}[!ht]
    \centering
    \subfigure{\includegraphics[width=1.0\textwidth]{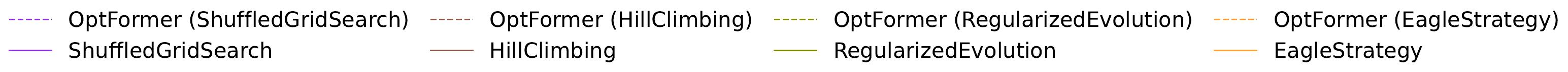}}\\
    \centering
    \subfigure{\includegraphics[width=0.19\textwidth]{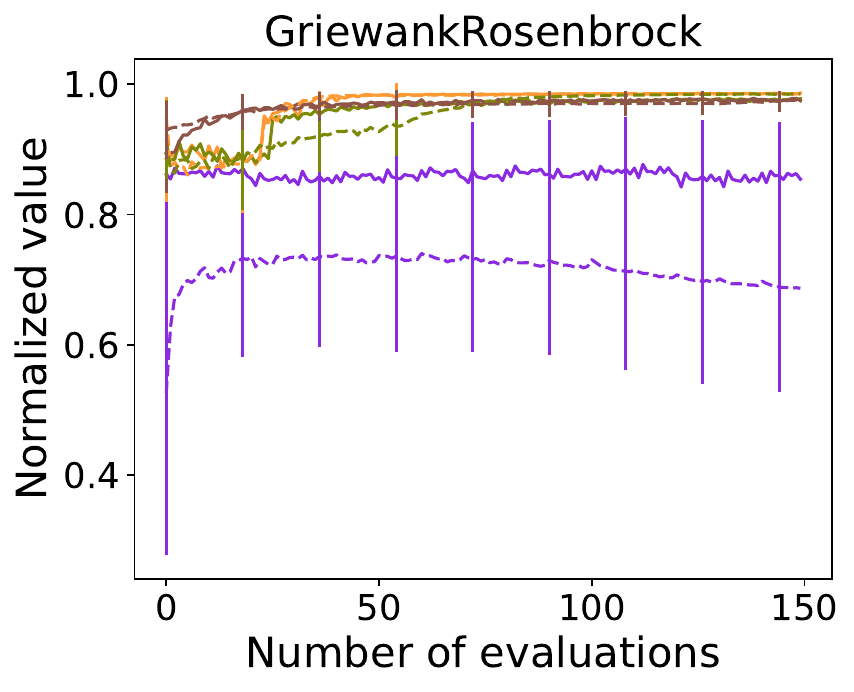}}
    \subfigure{\includegraphics[width=0.19\textwidth]{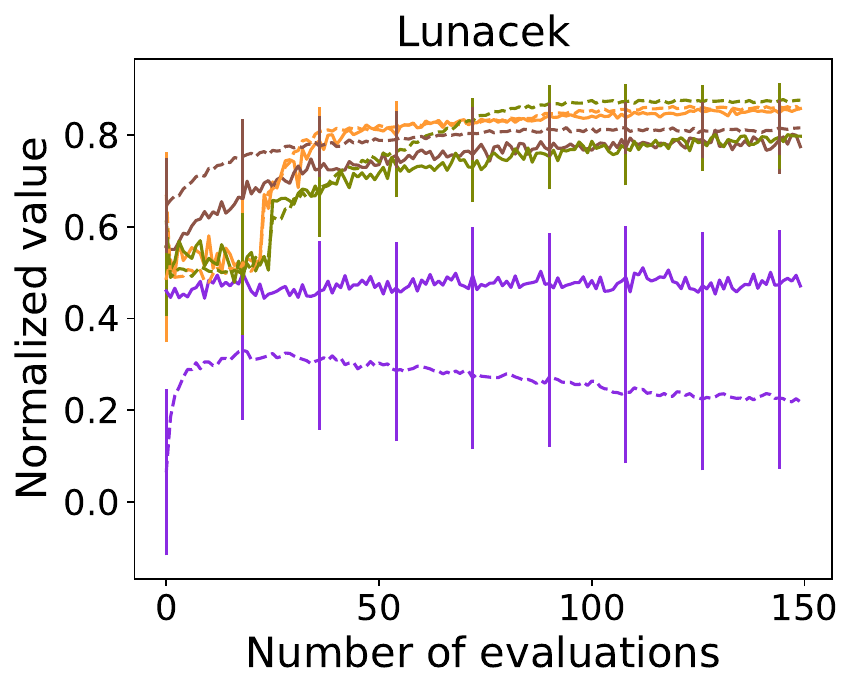}}
    \subfigure{\includegraphics[width=0.19\textwidth]{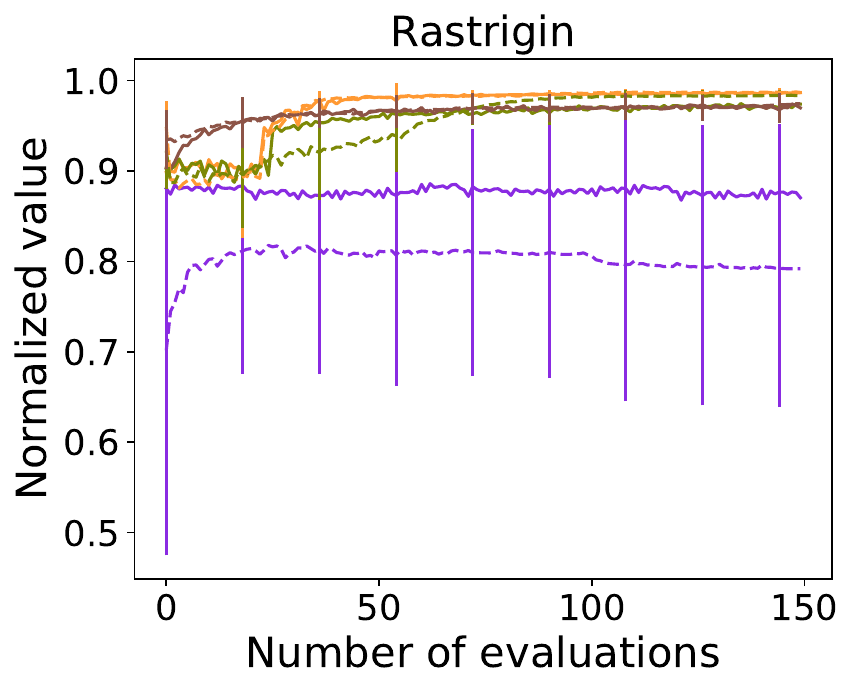}}
    \subfigure{\includegraphics[width=0.19\textwidth]{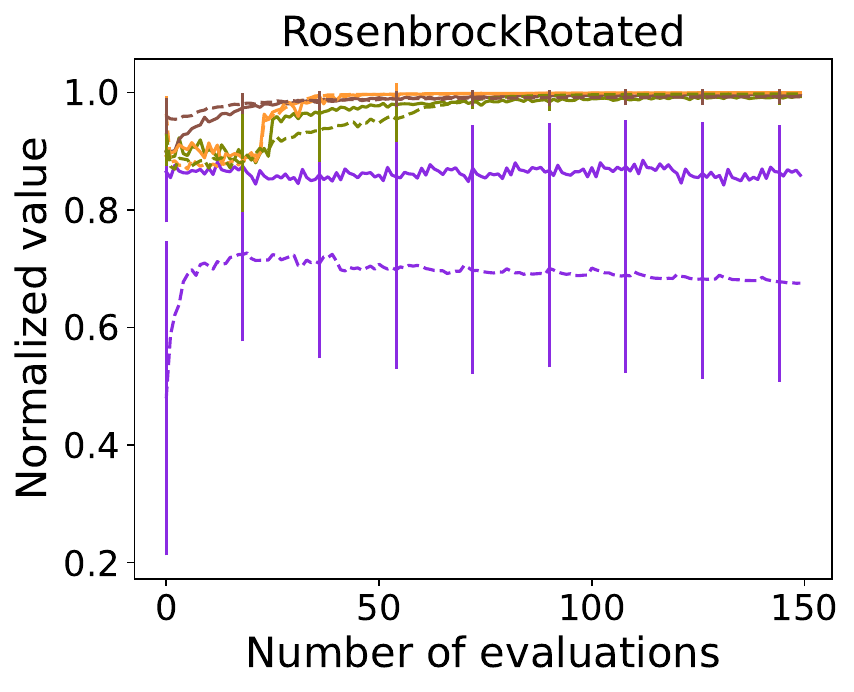}}
    \subfigure{\includegraphics[width=0.19\textwidth]{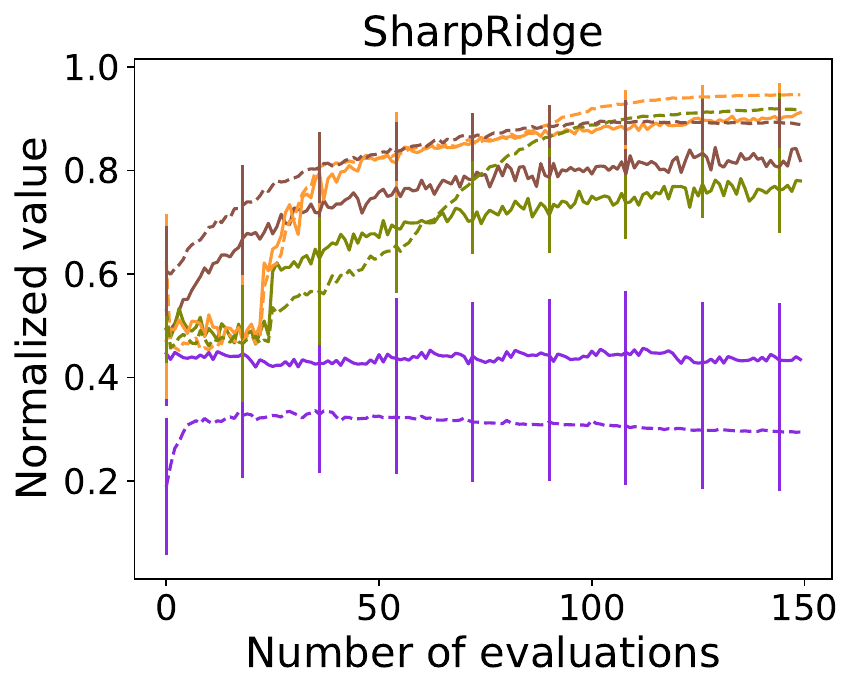}}
    \caption{Comparison of the behavior algorithms with the OptFormer re-implementation.}
    \label{fig:optformer}
\end{figure*}

To examine the algorithm imitation capability, we compare our re-implementation with the corresponding behavior algorithms. The results are shown in Figure~\ref{fig:optformer}. For clarity in the visual representation, we only plot a subset of the behavior algorithms, including Shuffled Grid Search, Hill Climbing, Regularized Evolution and Eagle Strategy. These behavior algorithms are plotted by solid lines, while their OptFormer counterparts are shown by dashed lines with the same color. Note that the figure uses the immediate function values as the $y$-axis to facilitate a comprehensive observation of the optimization process.

\section{Generalization}
\label{appendix:generalization}

\begin{figure*}[!ht]
    \centering
    \subfigure{\includegraphics[width=0.45\textwidth]{figures/main_exp/main_exp_legend1.pdf}}\\\vspace{-1em}
    \subfigure{\includegraphics[width=0.8\textwidth]{figures/main_exp/main_exp_legend2.pdf}}\\\vspace{-0.5em}
    \centering
    \subfigure{\includegraphics[width=0.19\textwidth]{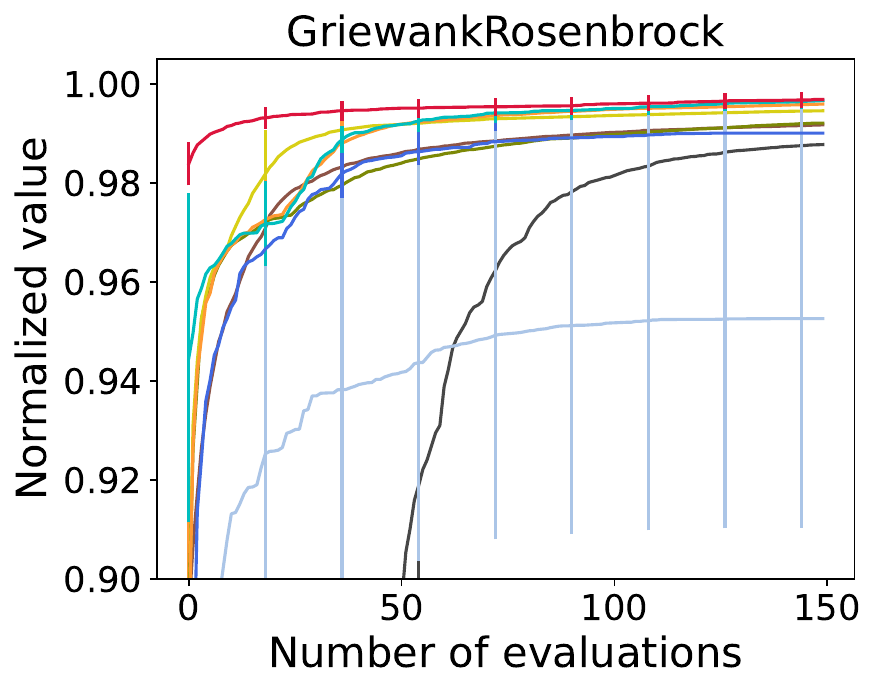}}
    \subfigure{\includegraphics[width=0.19\textwidth]{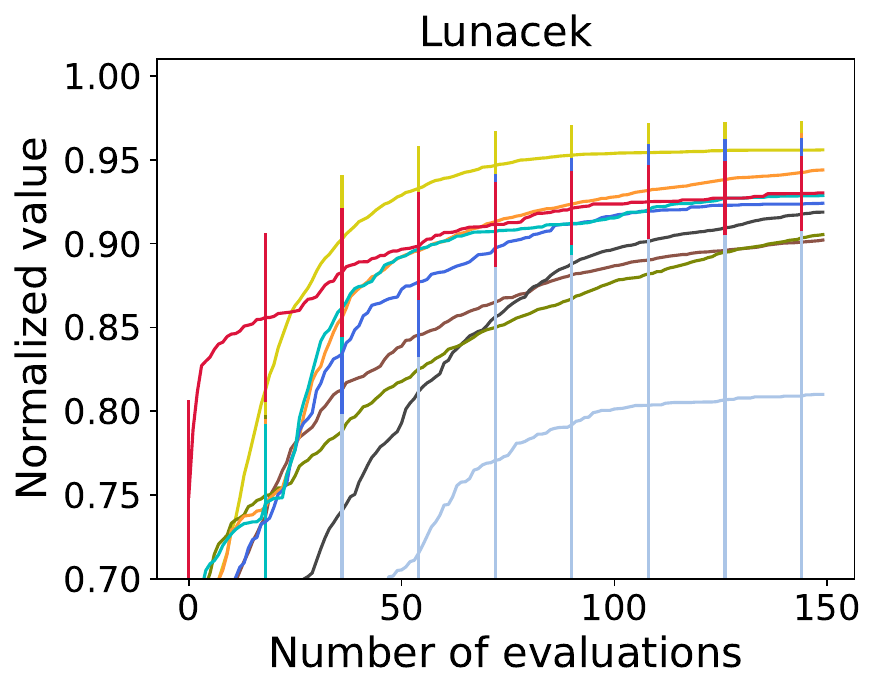}}
    \subfigure{\includegraphics[width=0.19\textwidth]{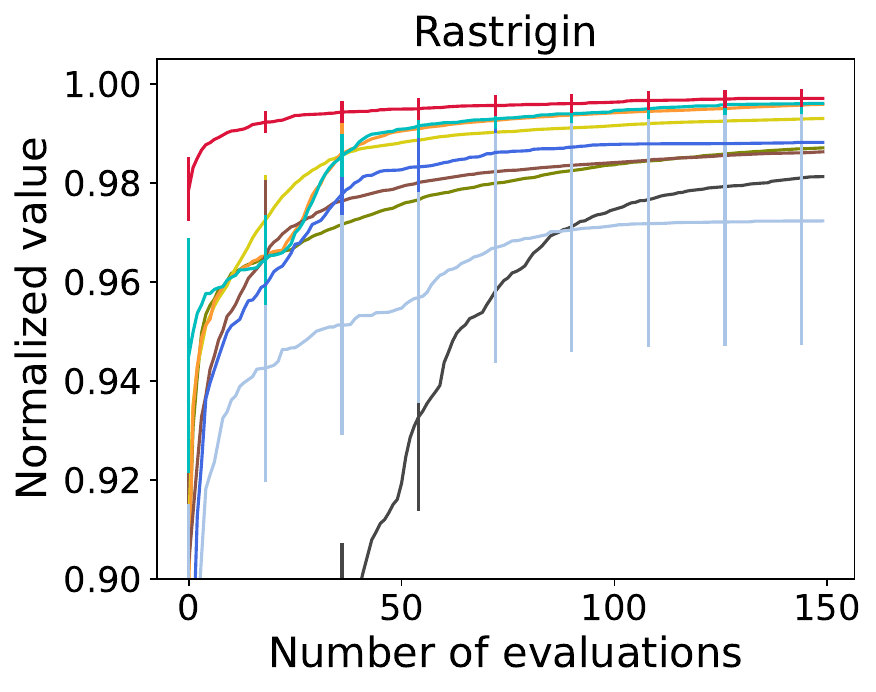}}
    \subfigure{\includegraphics[width=0.19\textwidth]{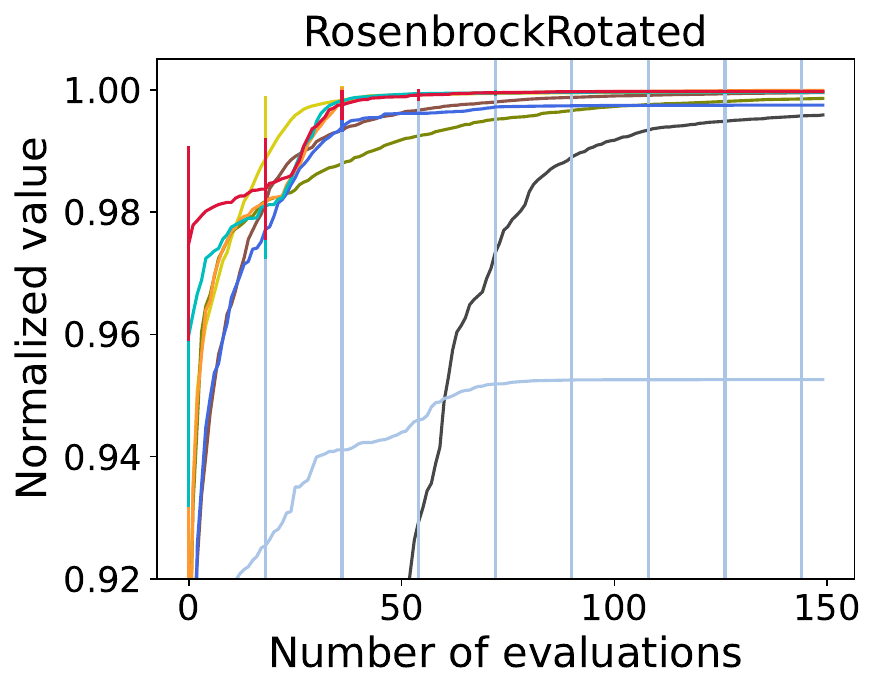}}
    \subfigure{\includegraphics[width=0.19\textwidth]{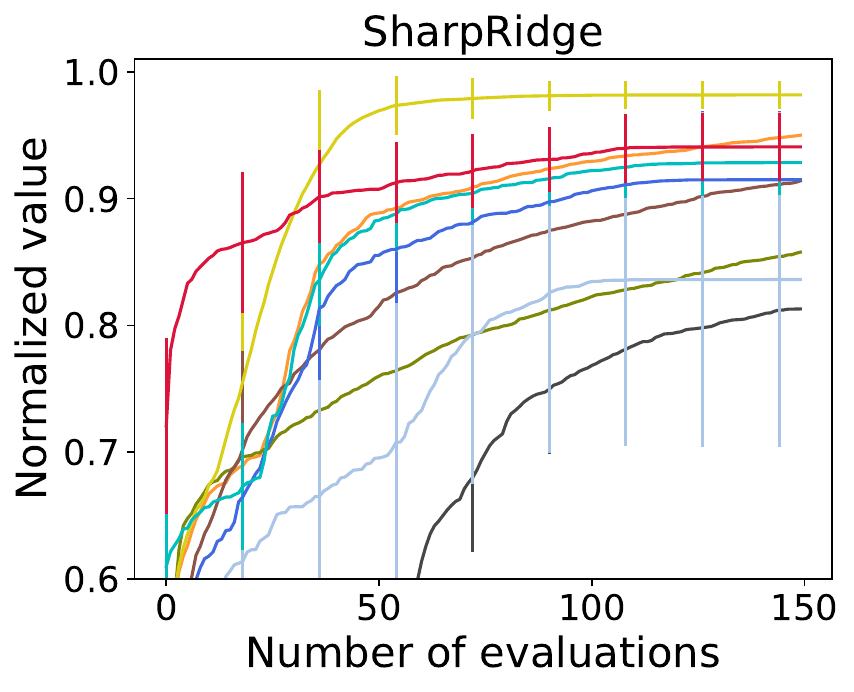}}
    \caption{Cross-distribution generalization by training on $4$ of $5$ chosen synthetic functions and testing on the remaining one.}
    \label{appendix:fig:generalization}
\end{figure*}

We conduct a series of experiments to examine the generalization of our method in this section. We train the model on $4$ of $5$ chosen synthetic functions and test on the remaining one. The results in Figure~\ref{appendix:fig:generalization} have shown that RIBBO has a strong generalization ability to unseen function distributions.

\begin{figure*}[!t]
    \centering
    \subfigure[Aggregation on the entire BBOB]{\label{appendix:fig:agg-bbob}\includegraphics[width=0.35\textwidth]{figures/generalization/average_bbob.pdf}}
    \subfigure[HRR with different $R_t$]{\label{appendix:fig:hrr}\includegraphics[width=0.35\textwidth]{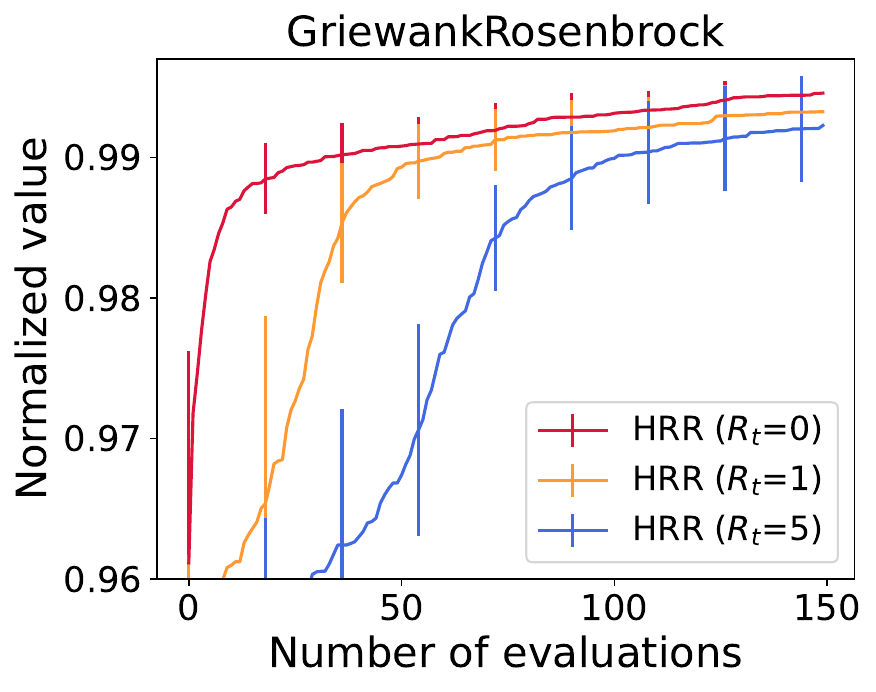}}
    \vspace{-0.7em}
\end{figure*}

We also train the model across all training datasets from BBOB synthetic functions simultaneously, and normalize all the test functions to aggregate results across functions with different output scaling. The results are shown in Figure~\ref{appendix:fig:agg-bbob}, and RIBBO has the best performance. 

\section{Further Discussions about RTG Tokens}
\label{appendix:rtg-diss}

We conduct more discussion about the setting of RTG tokens in this section. The immediate RTG tokens are designed to represent the performance based on cumulative regret over the future trajectories (i.e., $R_{i}=\sum_{t'=i+1}^t (y^* - y_{t'})$), rather than focusing solely on instantaneous regret $y^*-y_{t+1}$. Thus, setting the immediate RTG to $0$ inherently accounts for future regret, thereby enabling the algorithm to autonomously trade-off the exploration and exploitation. We conduct experiments to examine the performance of RIBBO with different values of $R_t$. The results are shown in Figure~\ref{appendix:fig:hrr}. Setting to a value larger than $0$ will decrease the performance and converge to a worse value, validating the effectiveness of setting the immediate RTG as $0$.

Sampling the next query point $\bm x_t$ is influenced by all the model inputs, including the immediate RTG token $\bm x_t$ and the previous history $(\bm x_i, y_i, R_i)^{t-1}_{i=0}$. The immediate RTG token represents the goal that is intended to be achieved, while the historical data contains information about the problem. These elements are integrated to construct the inputs, influencing the resulting sampled point. Even if the RTG token is set to $0$, implying a desire to sample the optimum, the short length of the history suggests limited knowledge about the problem, prompting the model to retain explorative behavior. As the history extends, suggesting a sufficient knowledge of the problem, the preference encoded by the RTG token shifts towards more greedy action. We conduct an experiment about the std of the output Gaussian head of $x$ of RIBBO using GriewankRosenbrock function and the results are shown in Table~\ref{tab-std}. It is clear that the std is very large in the early stage and becomes small later, which demonstrates the ability to trade-off the exploration and exploitation.

\begin{table}[!ht]
\centering
\begin{tabular}{|c|c|c|c|c|c|c|}
\hline
\textbf{Number of evaluations} & \textbf{1} & \textbf{30} & \textbf{60} & \textbf{90} & \textbf{120} & \textbf{150} \\ \hline
std of $\bm x$                 & 0.37       & 0.17        & 0.14        & 0.11        & 0.03         & 0.01         \\ \hline
\end{tabular}
\caption{Std of the output Gaussian head of $x$.}
\label{tab-std}
\end{table}


\section{Ablation Studies}
\label{appendix:ablation}

We provide further ablation studies to examine the influence of the key components and hyper-parameters in RIBBO, including the method to aggregate $(\bm x_i, y_i, R_i)$ tokens, the normalization method for the function value $y$, the model size, and the sampled subsequence length $\tau$ during training.

\begin{figure*}[!ht]
    \centering
    \subfigure[Token aggregation]{\label{fig:mix-method}\includegraphics[width=0.24\textwidth]{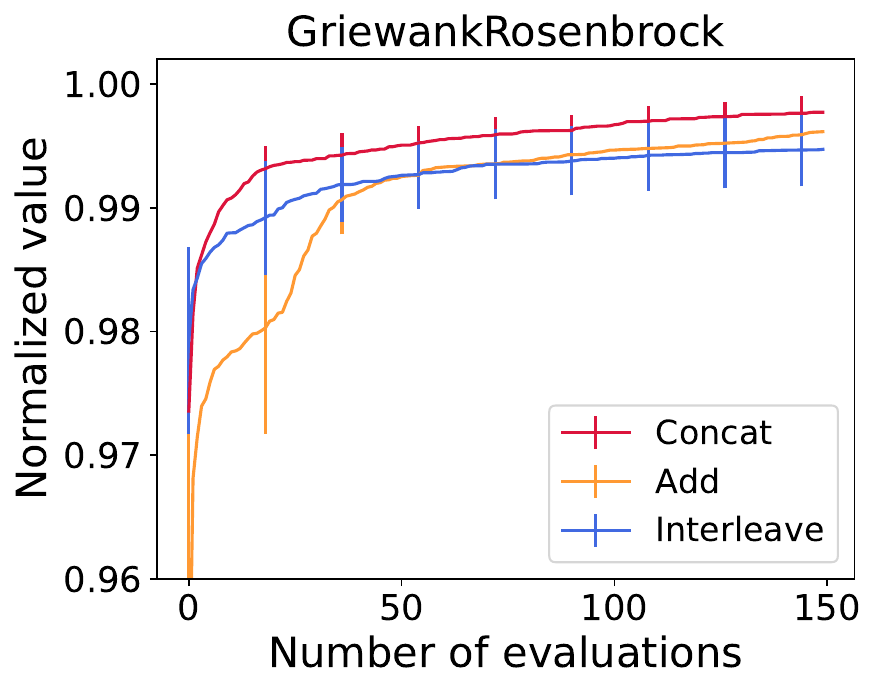}}
    \subfigure[Normalization method]{\label{fig:normalize-method}\includegraphics[width=0.24\textwidth]{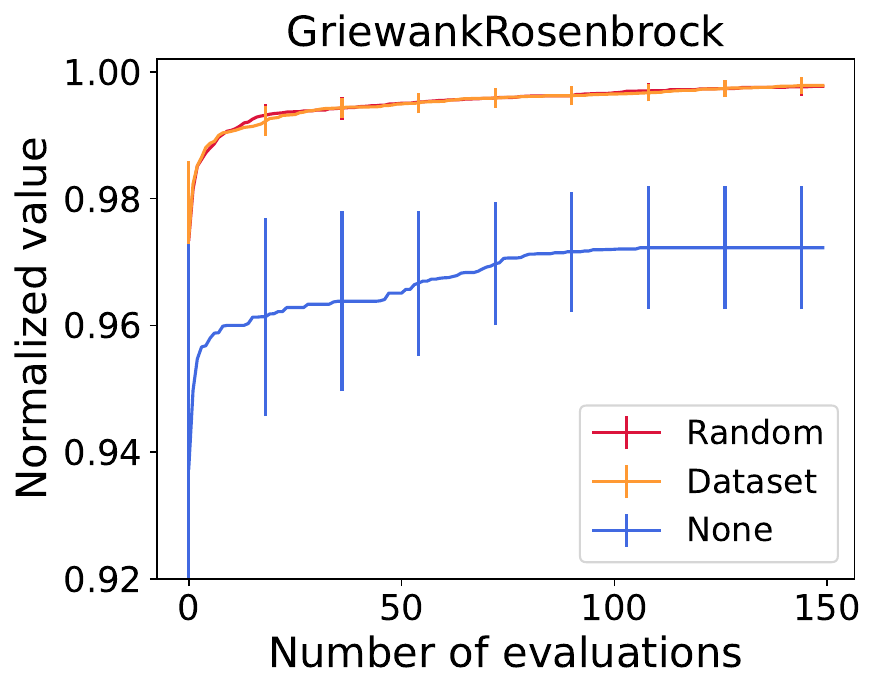}}
    \subfigure[Model size]{\label{fig:model-arch}\includegraphics[width=0.24\textwidth]{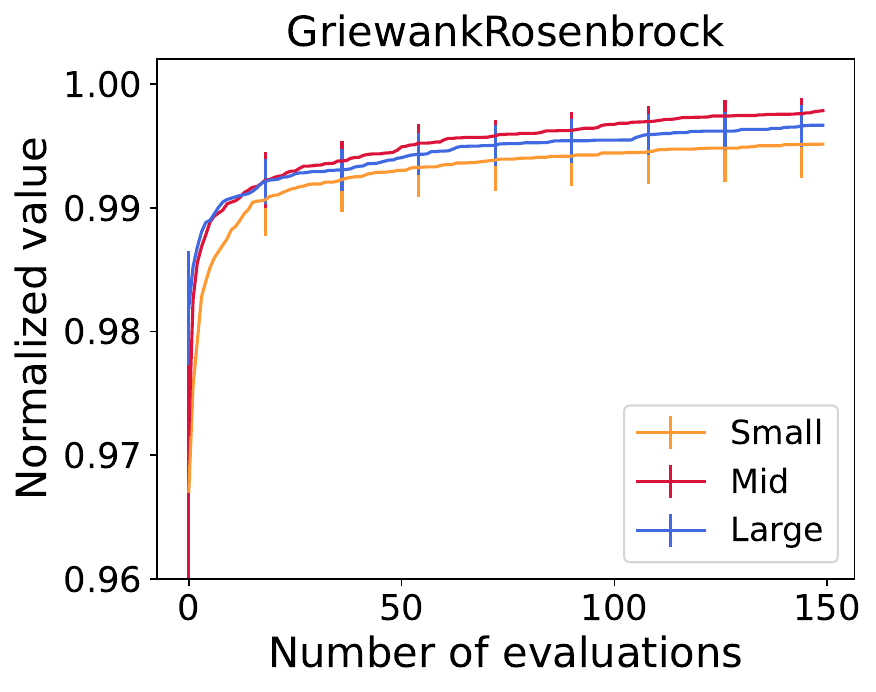}}
    \subfigure[Subsequence length $\tau$]{\label{fig:input-seq-len}\includegraphics[width=0.24\textwidth]{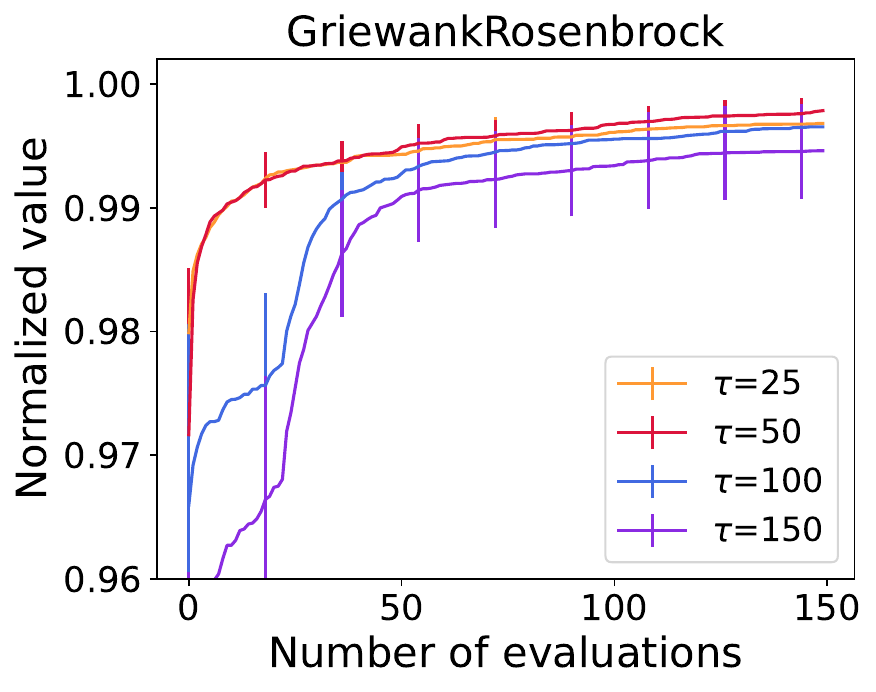}}
    \caption{Ablation studies of token aggregation, normalization method, model size, and subsequence length $\tau$.}
    \label{appendix:fig:ablation}
\end{figure*}

\textbf{Token Aggregation} aims to aggregate the information from $(\bm x_i, y_i, R_i)$ tokens and establish associations between them. RIBBO employs the Concatenation method (Concat), i.e., concatenating to aggregate $\bm x_i$, $y_i$ and $R_i$ to form a single token. This method is compared with two alternative token aggregation methods, i.e., Addition (Add) and Interleaving (Interleave). The addition method integrates the values of each token into one, while the interleaving method addresses tokens sequentially. The results are shown in Figure~\ref{fig:mix-method}. The concatenation method surpasses both the addition and interleaving methods, with the addition method showing the least efficacy. The concatenation method employs a relatively straightforward approach to aggregate tokens, while the interleaving method adopts a more complex process to learn the interrelations among tokens. The inferior performance of the addition method can be due to the summation operator, which potentially eliminates critical details from the original tokens. These details are essential for generating the subsequent query $\bm x_t$. 

\textbf{Normalization Method} is to balance the scales of the function value $y$ across tasks. We compare the employed random normalization with the dataset normalization and the absence of any normalization. Dataset normalization adjusts the value of $y$ by $(y - y_{\rm min}^i) / (y_{\rm max}^i - y_{\rm min}^i)$ for each task $i$, where $y_{\rm min}^i$ and $y_{\rm max}^i$ denote the observed minimum and maximum values of $f_i$. No normalization uses the original function values directly. The results in Figure~\ref{fig:normalize-method} indicate that random normalization and dataset normalization perform similarly, whereas the absence of normalization significantly hurts the performance. The choice of normalization method directly impacts the computation of RTG tokens and subsequently affects the generation of the desired regret. The lack of normalization leads to significant variations in the scale of $y$, which complicates the training process. Both random and dataset normalization scale the value of $y$ within a reasonable range, thus facilitating both training and inference. However, random normalization can bring additional benefits, such as invariance across various scales of $y$ as mentioned in~\citet{wistuba2021fewshot} and~\citet{chen2022towards}. Therefore, we recommend using random normalization in practice.

\textbf{Model Size} has an effect on the in-context learning ability. We assess the effects of different model sizes by comparing the performance of the currently employed model size with both a smaller and a larger model. The smaller model consists of $8$ layers, $4$ attention heads, and $128$-dimensional embedding space, while the larger model has $16$ layers, $12$ attention heads, and $384$-dimensional embedding space. The results are shown in Figure~\ref{fig:model-arch}, indicating that the model size has a minimal impact on overall performance. Specifically, the smaller model, due to its limited capacity, shows a reduced performance, while the larger one, potentially more powerful, requires more training data, which can lead to a slight decrease in performance due to overfitting or inefficiency in learning from a limited dataset. This analysis highlights the trade-offs involved in selecting the appropriate model size for optimal performance.

\textbf{Subsequence Length $\tau$} controls the context length during both the training and inference phases. The process of subsequence sampling acts as a form of data augmentation, enhancing training efficiency. As shown in Figure~\ref{fig:input-seq-len}, sampling subsequences rather than using the entire history as context, particularly when $\tau=T=150$, leads to improved performance. The computational complexity for causal transformer training and inference scales quadratically with the context length. Therefore, utilizing a shorter $\tau$ can significantly reduce computational demands. However, it is important to note that a shorter $\tau$ might not capture sufficient historical data, potentially degrading the performance due to the insufficient contextual information. This highlights the trade-offs between computational complexity and performance.

\section{Visualization of Branin Function}
\label{appendix:branin}

In several of our experiments, such as BBOB, rover problems, and the visualization analysis of the Branin function, we implement a series of transformations to the search space. This is designed to generate a distribution of functions with similar properties. A set of functions is sampled serving as the training and test tasks from the distribution. 

\begin{figure*}[!ht]
    \includegraphics[width=0.24\textwidth]{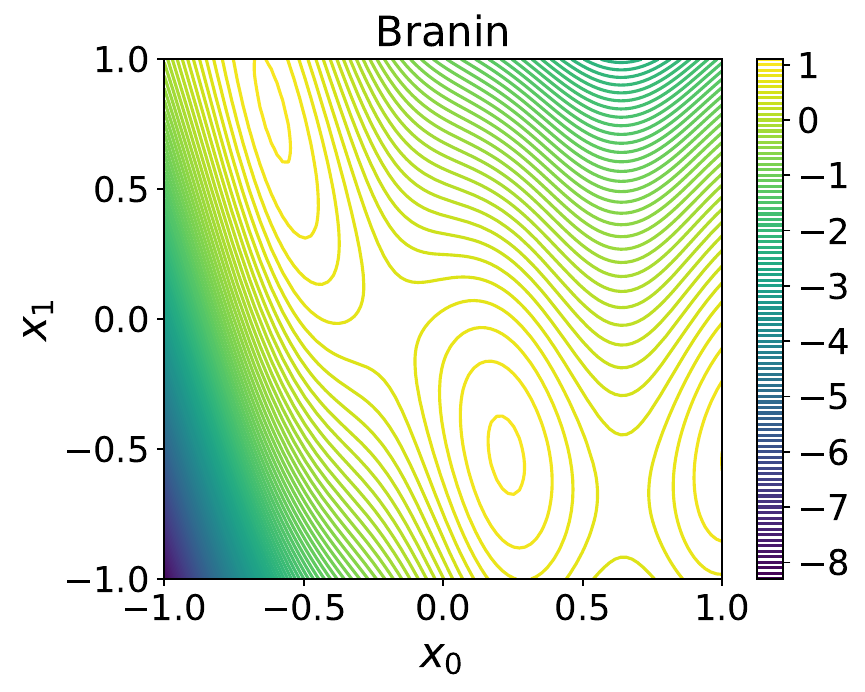}
    \includegraphics[width=0.25\textwidth]{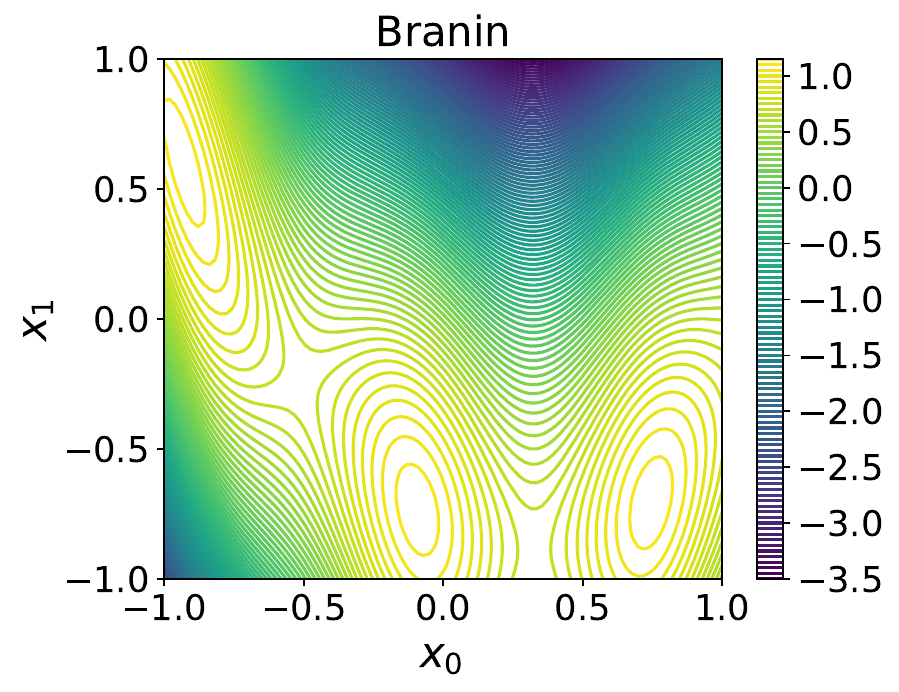}
    \includegraphics[width=0.24\textwidth]{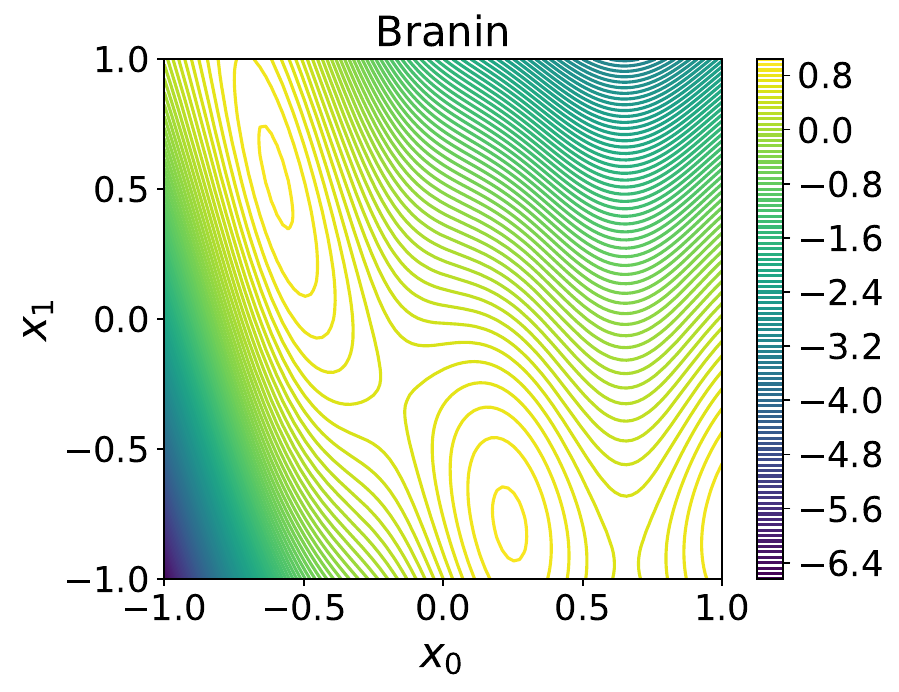}
    \includegraphics[width=0.24\textwidth]{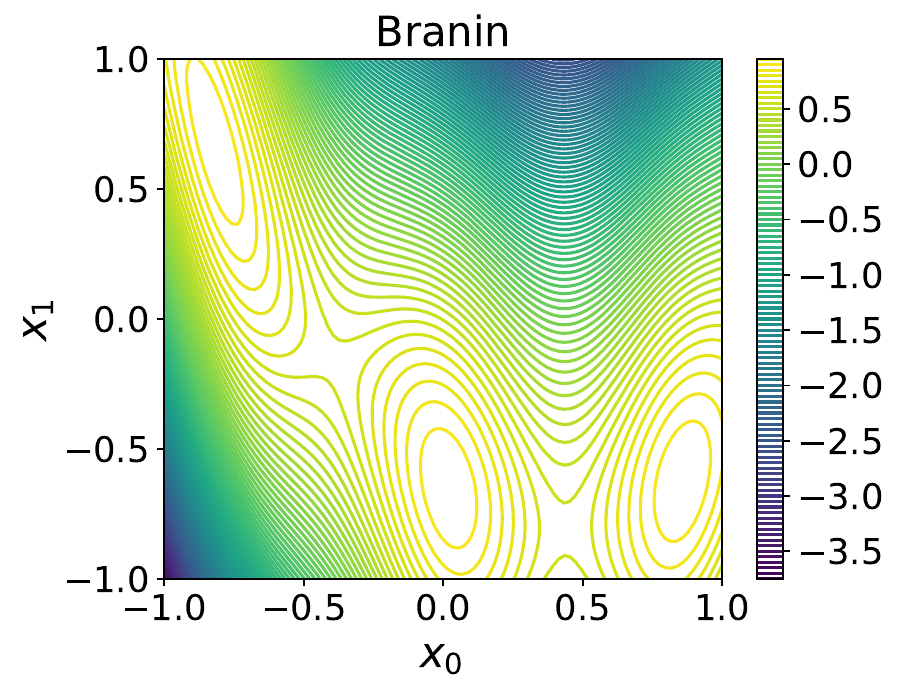}
    \caption{Contour line visualization for samples drawn from the $2$D Branin function distribution using $4$ distinct random seeds.}
    \label{appendix:fig:branin}
\end{figure*}

For the Branin function, random translations and scalings are applied to form the distribution. In this section, we present visualizations of the contour lines of the $2$D Branin functions, sampled from the distribution to demonstrate the effects of the applied transformations. To this end, $4$ distinct random seeds are used to draw samples from the distribution. The visualizations are shown in Figure~\ref{appendix:fig:branin}. Note that the two parameters of the Branin function have been scaled to the range of $[-1, 1]$ for the clarity of visual representations. Variations are observed in both the location of the optimum and the scales of the objective values. For instance, the optimum of the first subfigure locates more to the right compared to the second one, and the contour lines in the former are much more dispersed than those in the latter. The value scale for the first subfigure ranges from $-8$ to $1$, while in the second, it ranges from $-3.5$ to $1$. By applying these transformations, we can generate a function distribution that retains similar properties, enabling the sampling of training and test tasks for our model. 

For other functions, such as BBOB suite, besides simple random translations and scalings, more complex transformations such as non-linear Tosz and Tasy transformations are applied, leading to a more intricate landscape. However, due to the high dimensionality of these functions, direct visualization is impractical, so we present only the visualization of the $2$D Branin function. 

\end{document}